\pdfoutput=1
\documentclass[acmsmall]{acmart}

\AtBeginDocument{%
  }

\usepackage{algorithmic}
\usepackage{algorithm}
\usepackage{multirow}
\usepackage{xcolor}
\definecolor{lightgray}{RGB}{211, 211, 211}
\usepackage{enumitem}
\usepackage{verbatim} 

\usepackage{afterpage}
\usepackage{subcaption}
\usepackage{amsmath}
\usepackage{amsfonts}
\usepackage{amssymb}

\usepackage{graphicx}
\usepackage{caption}
\usepackage{float}
\usepackage{multicol}

\begin{document}

\title{Towards Multimodal Metaphor Understanding: A Chinese Dataset and Model for Metaphor Mapping Identification}



\author{Dongyu Zhang}
\email{zhangdongyu@dlut.edu.cn}
\affiliation{
\department{School of Foreign Languages}
  \institution{Dalian University of Technology}
  \city{Dalian}
  \postcode{116023}
  \country{China}
  }

\author{Shengcheng Yin}
\email{Cheng24@mail.dlut.edu.cn}
\author{Jingwei Yu}
\email{yjingwei@mail.dlut.edu.cn}
\affiliation{
\department{School of Software}
  \institution{Dalian University of Technology}
  \city{Dalian}
  \state{Liaoning}
  \postcode{116023}
  \country{China}
  }

\author{Zhiyao Wu}
\email{bc31072@um.edu.mo}
\affiliation{
\department{Faculty of Business Administration}
  \institution{University of Macau}
  \city{Macau}
  \postcode{519000}
  \country{China}
  }

\author{Zhen Li}
\email{lzhen0205@gmail.com}
\affiliation{
\department{Faculty of Business and Commerce}
  \institution{Kansai University}
  \city{Osaka}
  \postcode{564-8680}
  \country{Japan}
  }

\author{Chengpei Xu}
\email{Chengpei.Xu@unsw.edu.au}
\affiliation{
\department{School of Minerals and Energy Resources Engineering}
  \institution{University of New South Wales}
  \city{Sydney}
  \state{NSW}
  \postcode{2052}
  \country{Australia}
  }

\author{Xiaoxia Wang}
\email{spice.wang@rmit.edu.au}
\affiliation{
\department{Centre for Educational Innovation and Quality}
  \institution{RMIT University}
  \city{Melbourne}
  \state{VIC}
  \postcode{3000}
  \country{Australia}
  } 

\author{Feng Xia}
\authornote{Corresponding author}
\email{f.xia@ieee.org}
\affiliation{
\department{School of Computing Technologies}
  \institution{RMIT University}
  \city{Melbourne}
  \state{VIC}
  \postcode{3000}
  \country{Australia}
  }









\begin{abstract}
Metaphors play a crucial role in human communication, yet their comprehension remains a significant challenge for natural language processing (NLP) due to the cognitive complexity involved. According to Conceptual Metaphor Theory (CMT), metaphors map a target domain onto a source domain, and understanding this mapping is essential for grasping the nature of metaphors. While existing NLP research has focused on tasks like metaphor detection and sentiment analysis of metaphorical expressions, there has been limited attention to the intricate process of identifying the mappings between source and target domains. Moreover, non-English multimodal metaphor resources remain largely neglected in the literature, hindering a deeper understanding of the key elements involved in metaphor interpretation. To address this gap, we developed a Chinese multimodal metaphor advertisement dataset (namely CM3D) that includes annotations of specific target and source domains. This dataset aims to foster further research into metaphor comprehension, particularly in non-English languages. Furthermore, we propose a Chain-of-Thought (CoT) Prompting-based Metaphor Mapping Identification Model (CPMMIM), which simulates the human cognitive process for identifying these mappings. Drawing inspiration from CoT reasoning and Bi-Level Optimization (BLO), we treat the task as a hierarchical identification problem, enabling more accurate and interpretable metaphor mapping. Our experimental results demonstrate the effectiveness of CPMMIM, highlighting its potential for advancing metaphor comprehension in NLP. Our dataset and code are both publicly available to encourage further advancements in this field.
\end{abstract}





\maketitle

\section{Introduction}

Metaphor plays a pivotal role in human cognition and communication, appearing approximately once every three sentences in everyday language  \cite{steen2010method}. In Natural Language Processing (NLP) and Information Retrieval (IR), unraveling the intricacies of metaphors has become a significant challenge. The Conceptual Metaphor Theory suggests that metaphors involves mapping a target domain to a source domain \cite{lakoff1980metaphors}. To attain a profound comprehension of metaphors, it becomes imperative to identify both domains.

\begin{figure}[htb]
    \centering
    \begin{minipage}[b]{0.36\linewidth}
        \centering
        \includegraphics[height=6cm]{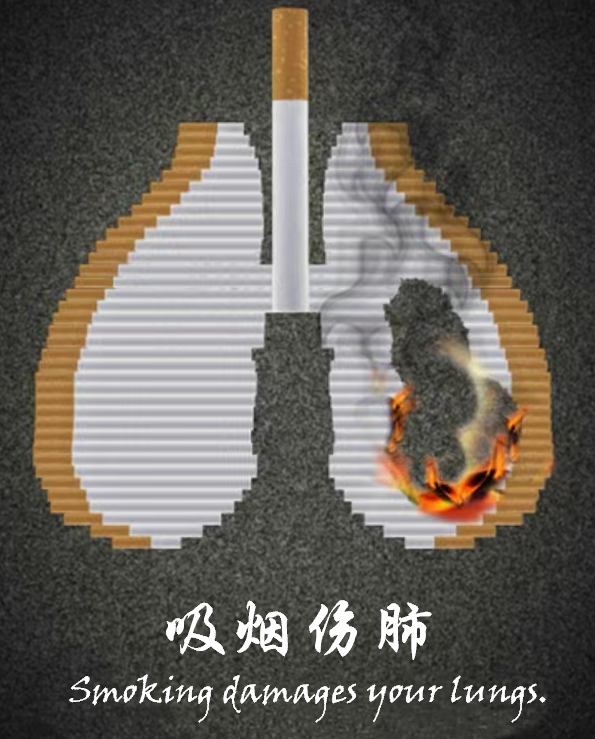}
        \caption{Example of multimodal metaphor.}
        \label{fig:figure_1}
    \end{minipage}
    \hspace{0.1cm}
    \begin{minipage}[b]{0.54\linewidth}
        \centering
        \includegraphics[height=6cm]{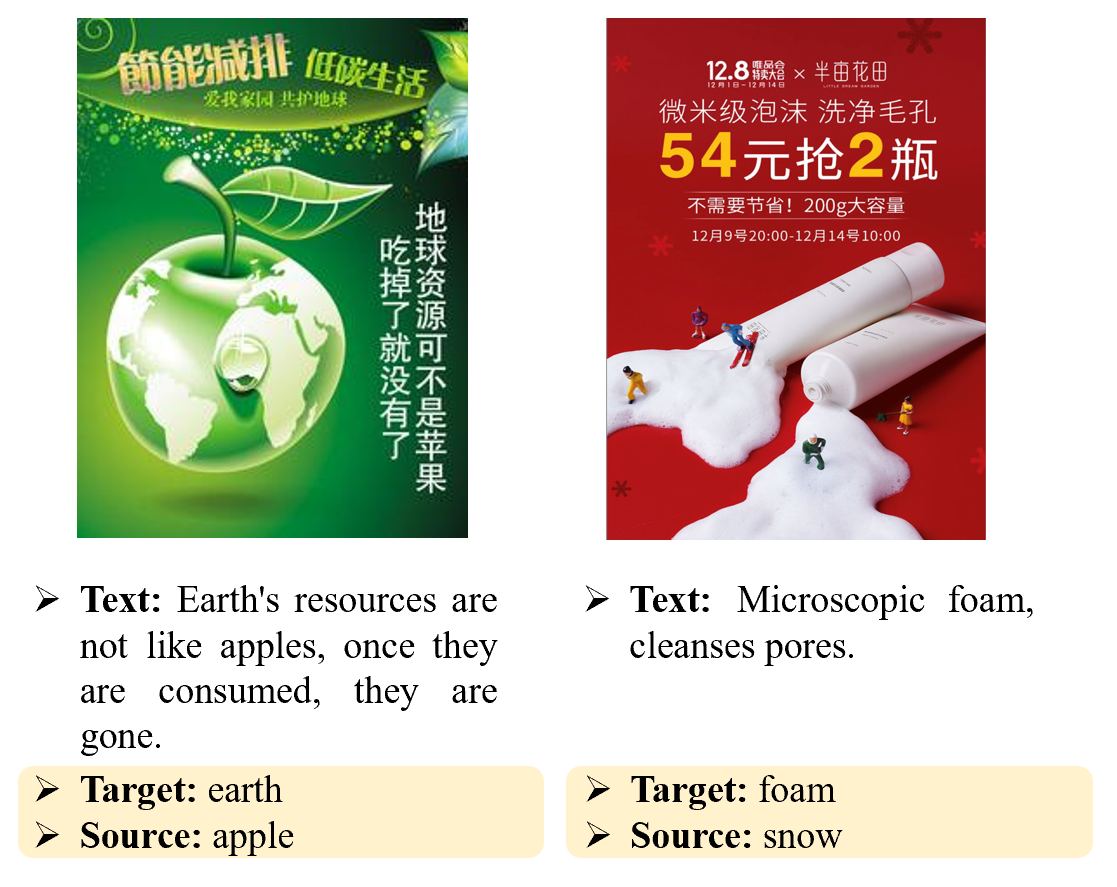}
        \caption{Annotation examples.}
        \label{fig:figure_3}
    \end{minipage}
\end{figure}

With the advent of modern media, multimodal metaphors has significantly increased, surpassing monomodal counterparts due to their vivid, attractive, and persuasive effects \cite{xubomm2024,xubosigir22}. This is especially true in fields like advertising, marketing, and recommendation \cite{6594800}. A multimodal metaphor, as articulated by Forceville et al. \cite{forceville2009multimodal}, involves a mapping that conceptualizes one target domain in terms of another source domain, utilizing different modes such as text and image, text and sound, or image and sound. Figure \ref{fig:figure_1} illustrates a compelling example:lungs constructed from cigarettes to symbolize that smoking causes lung damage.  This visual metaphor symbolically connects two distinct entities—the \textit{lung} and the \textit{cigarette}—evoking the perceptual notion that smoking is a primary cause of lung damage. The \textit{cigarette} image, representing the source domain, intricately intertwines with the target domain, depicted both textually and visually through the representation of the \textit{lung}.

The understanding of multimodal metaphors primarily centers on identifying the underlying mapping between these two domains, offering valuable insights into the cognitive mechanisms that shape metaphorical thinking. By pinpointing these domains, researchers gain a deeper understanding of the processes facilitating the transfer of concepts from one domain to another. The identification of target and source domains in metaphor analysis is of paramount significance, unlocking profound insights into the cognitive processes involved in metaphor interpretation.

In NLP, previous research has predominantly centered on tasks such as detecting whether a phrase is metaphorical or literal \cite{kuo-carpuat-2020-evaluating,liu-etal-2020-metaphor,choi-etal-2021-melbert} and and analyzing the sentiments conveyed by metaphors  \cite{Yang2013ContextualEO}, particularly in the English language. However, most studies have concentrated on English texts and often overlook the critical task of identifying target and source domains. Additionally, research on multimodal metaphors in non-English languages remains scarce, leaving a significant gap in the field. 

To address these gaps, this paper introduces the Chinese Multimodal Metaphor Mapping Dataset (CM3D). This dataset is meticulously curated to encompass annotations of metaphorical expressions within both target and source domains, with a specific focus on the Chinese language. Additionally, 
we propose the CoT Prompting-based Metaphor Mapping Identification Model (CPMMIM). This model leverages the concept of CoT to enhance the recognition of target and source domains in metaphors, offering a fresh perspective on metaphor understanding. Experimental results demonstrate the effectiveness of this model in accurately identifying target and source domains. Our contributions in this research can be outlined as follows: 
\begin{enumerate}
\item \textbf{The CM3D Dataset}:  a  collection of 6,108 text-image pairs sourced from Chinese advertisements. This dataset includes meticulous annotations of metaphorical expressions within target and source domains, specifically tailored for the Chinese language.

\item \textbf{The CPMMIM Mode}l: A novel method for metaphor mapping identification based on Bi-Level Optimization and Chain-of-Thought prompting, which mirrors cognitive reasoning processes to extract target and source domains from metaphors. By leveraging this approach, our model can extract specific words within both target and source domains in metaphorical instances, fostering a deeper understanding of metaphors.

\item \textbf{Evaluation Benchmark: }The CM3D dataset serves as a benchmark for testing various baseline models. Through extensive experiments and analysis, we evaluate the performance of these models in recognizing multimodal metaphors.
\end{enumerate}

These contributions address key gaps in existing research on metaphor datasets and the extraction of target and source domains. By providing a dedicated resource and innovative approach for Chinese multimodal metaphors, this work lays a foundation for further exploration and benchmarking in this field.

\section{Related work}

\subsection{Multimodal Metaphor Datasets}

The exploration of multimodal metaphor datasets is still in its eqarly stages, with only a limited number currently available. While various textual metaphor datasets have been introduced for metaphor processing in NLP, such as those presented by Steen et al. \cite{steen2010method}, Mohammad et al. \cite{mohammad-etal-2016-metaphor}, Rosen \cite{rosen2018computationally} and Wachowiak et al. \cite{wachowiak-gromann-2023-gpt}, the development of multimodal metaphor datasets has been relatively sparse. Notably, researchers like Shutova et al. \cite{shutova2017multilingual} and Zhang et al. \cite{zhang-etal-2021-multimet}, who constructed multimodal samples to investigate metaphor processing across text and other modalities. 

However, existing multimodal datasets, including those by Shutova et al. \cite{shutova2017multilingual} and Zhang et al. \cite{zhang-etal-2021-multimet}, have several limitations. First, they consist exclusively of English samples, restricting their utility for cross-linguistic studies. Second, these datasets primarily focus on metaphor detection—distinguishing between metaphorical and literal expressions—without exploring the deeper mechanisms underlying multimodal metaphor development.  In contrast, our dataset addresses these gaps by including Chinese samples, thereby contributing to linguistic diversity. Additionally, it provides detailed annotations that elucidate the mechanisms of metaphor creation and interpretation. These distinctive features make our dataset a valuable resource for researchers seeking to advance the understanding of multimodal metaphor comprehension and the intricate processes through which metaphorical expressions emerge.

\subsection{Extraction of Source Domain and Target Domain}

Among the existing work related to metaphor, the extraction of source and target domains is not new. Previous methods for extracting source or target domains often relied on dictionaries or datasets with specific syntactic structures tailored to identify the source domain when the target domain is given. For instance, Mohler et al. \cite{mohler2016introducing} introduced a metaphorical dataset containing the intensity of word pairs, encompassing 114 annotated source conceptual domains and 32 target conceptual domains. Building upon Mohler et al.'s dataset, Rosen et al. \cite{rosen2018computationally} combined deep neural network models with syntactic structure features and contextual information too predict the source domain associated with a given target domain. Dodge et al. \cite{dodge2015metanet} introduced a system capable of automatically detecting, classifying, and analyzing metaphors within a corpus, providing support for deep semantic interpretation based on these analysis results. Mao et al. \cite{mao2022metapro} presented an end-to-end English metaphor processing model that excels in identifying token-level metaphors within input text fields, offering natural language interpretations and explaining multi-word expressions of metaphors. While their method for detecting Multi-Word Expressions (MWE), which is based on dictionaries and rules, achieves notable coverage and accuracy, it is constrained by the size of the corpus and lacks broad generalization.

Some recent work has established metaphorical mappings to assist in the identification of source and target domains. Shutova et al. \cite{shutova2017multilingual} employed various clustering strategies to explore the relationship between source concept domains and target concept domain clusters under semi-supervised and unsupervised conditions. Ge et al. \cite{ge2022explainable} proposed a word-pair level metaphor detection system that identifies and interprets source-target domain word pairs by learning the hypernym relations in WordNET and constructing conceptual mappings. In more recent research, Wachowiak et al. \cite{wachowiak-gromann-2023-gpt} examined GPT-3's ability to comprehend metaphorical language without predefined domains.  They assessed the model's accuracy in predicting source domains across languages, fine-tuning, and applying few-shot learning with diverse training samples on two datasets. Su et al. \cite{su2024efficient}generated metaphor interpretation text based on the target domain and source domain in metaphorical sentences, combined the enhanced prompt information from the source domain with item image generation into a complete prompt, and applied it to the metaphorical image generation task.

We consider the limitations of previous work and make improvements in our work:  Our study introduces the task of Chinese metaphorical relation extraction for the first time, extends metaphorical mapping extraction to the multimodal level, and applies large models to the extraction of target and source domains. Our approach addresses resource limitations in corpora, enhances generalization, and aims to contribute to future developments in related fields.

\subsection{Chain of Thought Prompting on Large Language Models}

Previous research has mainly employed fine-tuned pretrained models to handle metaphor-related tasks \cite{leong-etal-2020-report}, but these models have not fully utilized common sense and background information, leading to issues of misjudgment. However, the emergence of Large Language Models (LLMs) has changed this situation. LLMs are trained on extensive and diverse text corpora, enabling them to learn rich patterns of associations between words. Previous studies have shown that LLMs like GPT-3 perform well on complex tasks, such as code summarization \cite{10.1145/3551349.3559555}, commonsense understanding \cite{paranjape-etal-2021-prompting,liu-etal-2022-testing}, mathematical reasoning \cite{10534945, 10.5555/3600270.3602070,DBLP:conf/aaai/WangHHXLLS24,Cohn2024ACP}, relation extraction \cite{10337264}, and more. Therefore, leveraging the common sense and background information provided by LLMs can enhance the expression and comprehension of metaphors.

Recent research has shown that when language models are prompted to generate intermediate reasoning steps before answering questions, their performance is significantly better than directly providing answers \cite{DBLP:journals/corr/abs-2112-00114, wei2022chain, NEURIPS2022_8bb0d291}. prompting effectively reduces the hallucination phenomenon in text generation of LLMs \cite{ji2024chain}. Wei et al. \cite{wei2022chain} formally studied the Chain of Thought (CoT) promoting approach in language models, primarily by manually designing examples to assist in generating reasoning paths. Hao et al. \cite{fei-etal-2023-reasoning} and Gu et al. \cite{GU2024107907} applied the Chain of Thought to emotional inference, significantly improving the accuracy of emotion recognition.
In the multimodal domain, Zhang et al. \cite{zhang2023multicot}, Zheng et al. \cite{zheng-etal-2024-enhancing-semantics}, and He et al. \cite{he2024multi} have all employed the CoT approach to address multimodal question-answering problems, and achieved remarkable results.
 
Previous research has shown that in some classic experiments in cognitive psychology, LLMs can exhibit similar behavior to humans \cite{Binz_2023} and demonstrate human-like content effects in logical reasoning tasks \cite{Dasgupta2022LanguageMS}. Therefore, it is plausible that LLMs can establish word associations similar to humans, enabling metaphor understanding.When people hear a metaphor, they are not only comprehending its meaning but also attempting to deduce additional information, such as the true nature of the entity being described. This process of inference can be modeled and analyzed using cognitive probability models. For instance, Kao et al. \cite{Kao2014NonliteralUO} proposed a model that accurately predicts humans' inferences about the literal attributes of metaphorical references. Prystawski et al. \cite{prystawski2023psychologicallyinformed} connected CoT with the large-scale language model GPT-3, prompting the model to generate explanations involving the identification of implicit variables and describing the relationships between these variables, and then selecting the most suitable interpretation for the metaphor. 

Prystawski et al. \cite{prystawski2023psychologicallyinformed} demonstrated that if chain of thought prompts guide language models through processes resembling human reasoning, they may be particularly useful for understanding metaphors. However, this work only dealt with a single modality, and it was shown that relying solely on the power of large models for metaphor interpretation can lead to poor performance on highly novel metaphors. In contrast, we establish connections between LLMs and small-scale models, utilizing prompt-based LLMs to provide related knowledge and simulate the thinking process of human metaphor comprehension. Through training, we enable the model to learn novel metaphors and identify important mapping relationships in multimodal metaphors.

\subsection{Bi-Level Optimization}

Bi-Level Optimization (BLO) originates from the area of economic game theory and was then introduced into the optimization community. BLO is able to handle problems with a hierarchical structure, involving two levels of optimization tasks, where one task is
nested inside the other \cite{liu2021investigating}. The inner (or nested) and outer optimization tasks are often respectively referred to as the Lower-Level (LL) and Upper-Level (UL) subproblems \cite{dempe2020bilevel}. A basic BLO problem can be formalized as:
\begin{equation}
\min_{x \in X} F \left( x, y^* \right), \quad \text{s.t.} \quad y^* \in \min_{y \in Y } f \left( x, y \right)
\end{equation}
where \( F \) and \( f \) are the objective functions of the Upper-Level and Lower-Level subproblems, respectively; \( x \) is the decision variable of the Upper-Level subproblem; \( y \) is the decision variable of the Lower-Level subproblem; and \( y^* \) is the optimal solution obtained by solving the Lower-Level problem given \( x \). The above BLO problem has a natural interpretation as a non-cooperative game between two players \cite{dempe2020bilevel}. To illustrate this intuitively, consider the example of a dealer and a producer. The dealer's goal is to maximize profit, which depends on the production cost and market price of the product. Meanwhile, the producer's goal is to minimize production costs, which depend on product prices and production parameters. So this BLO problem can alse be formalized as:
\begin{equation}
\max_{P} Profit \left( p, c^* \right), \quad \text{s.t.} \quad c^* \in \min_{c} cost \left( p, c \right)
\end{equation}
where \( p \) is the product price and \( c \) is the production cost. Dealers first choose strategies with the goal of maximizing profits. This influences producers to adjust their production strategies and optimize their production costs. In turn, producers' decisions affect the dealers' profits.

BLO problems are commonly encountered in the field of computer vision, including complex problems such as hyper-parameter optimization, multi-task and meta-learning, and adversarial learning. Liu et al. \cite{liu2022target} proposed a BLO framework for joint fusion and detection tasks, aimed at enhancing the fusion of infrared and visible images by accounting for modal differences and capturing complementary information. Additionally, Liu et al. \cite{liu2022task} introduced a new framework named Task-Oriented Latent Feasibility(TOLF), which involves solving convex bi-level formulations to achieve optimal solutions for specific tasks. In this article, we consider our task to be a hierarchical problem, where the identification of the source domain can be viewed as an Upper-Level sub-task. Therefore, we draw on the concept of BLO to approach the problem.

\section{CM3D Dataset}

\begin{figure}[H]
	 	\centering  
	 	\includegraphics[width=\textwidth]{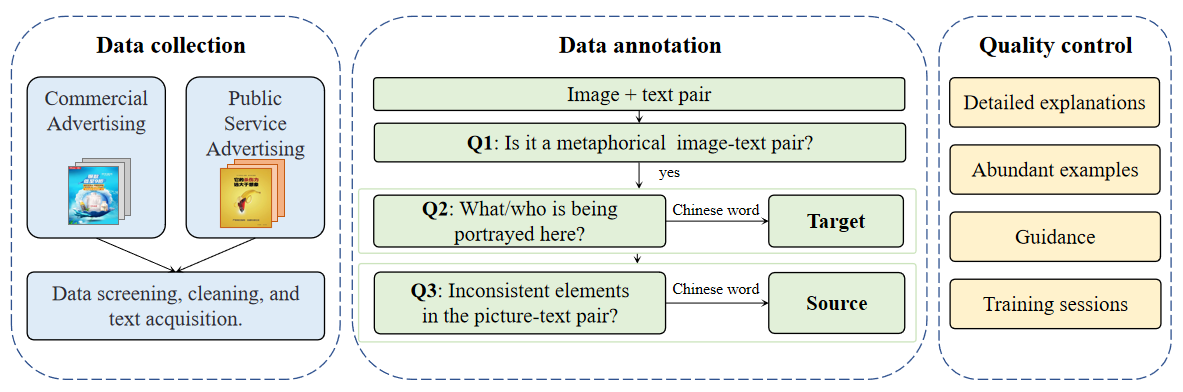}    
	 	\caption{Dataset Construction.}
	 	\label{fig:figure_2}
	 \end{figure}

Our dataset construction process is divided into three main stages: data collection, data annotation, and quality control. The overall workflow is illustrated in Figure \ref{fig:figure_2}. Below, we provide a detailed explanation of each stage.

\subsection{Data Collection}

We have chosen advertisements that integrate both textual and visual elements as our primary data source due to their pivotal role in multi-modal metaphor research, as underscored by Forceville \cite{forceville2017visual}. Whether serving commercial or public service objectives, these prevalent advertisement formats possess an intrinsic appeal as effective communication tools. They encapsulate a vast reservoir of metaphorical information complemented by a diverse array of visual and linguistic features. This decision is motivated by the recognition of their significance in enabling nuanced analyses and insights into the complexities of multi-modal metaphor usage.

Specifically, we curated examples of Chinese multi-modal metaphors from a large publicly available dataset \cite{zhang-etal-2023-multicmet}. This dataset comprises 13,820 text-image pairs of advertisements with manual annotations indicating the presence of metaphors. The data was procured through searches using Chinese keywords on platforms such as Baidu and Bing, as well as through participation in the IFlytek Advertising Image Classification competition.
competition\footnote{\url{https://aistudio.baidu.com/aistudio/datasetdetail/102279}}.

During the data filtering stage, we first selected the text-image pairs that were annotated as metaphors. Then, according to the filtering criteria, we retained pairs with clear text and dimensions larger than 350$*$350 pixels, while excluding PPT-style images that only contained text. As a result, we obtained 6,108 text-image pairs. Detailed statistics can be found in Table \ref{tab:table_1}.

\begin{table}[!ht]
    \normalsize
    \centering
     \caption{Statistics of the dataset.}
    \begin{tabular}{cccccc}
    \toprule[1pt]
        ~ & \textbf{Total Samples} & \textbf{Target Words} & \textbf{Source Words} & \textbf{Total Words} & \textbf{Avg Words} \\ 
    \midrule[0.5pt]
        Train & 4888 & 14764 & 10016 & 53,768 & 11 \\ 
        \addlinespace[0.3pt]
        Val & 610 & 1832 & 1212 & 6405 & 10 \\ 
        \addlinespace[0.3pt]
        Test & 610 & 1834 & 1215 & 7930 & 13 \\ 
    \midrule[0.5pt]
        Total & 6,108 & 18428 & 12443 & 68103 & 11 \\ 
    \bottomrule[1pt]
    \end{tabular}
   
    \label{tab:table_1}
\end{table}

\subsection{Domain Annotation}

Our annotation task focuses on identifying specific target and source domains within metaphorical image-text pairs. Following the method proposed by \cite{zhang-etal-2021-multimet}, annotators identify metaphorical image-text pairs by observing incompatible elements and interpreting the irreversibility of the \textit{A\ is\ B} identity relationship. In this dataset, the annotators are tasked with identifying specific \(A\) and \(B\) and expressing them in concise Chinese words. The annotation process involves a thorough analysis of the entire text+image pair, encompassing both visual and linguistic elements. The annotators begin by posing the question, "What/who is being portrayed here?" in order to ascertain the target domain within the metaphorical context. If the intention behind the image is ambiguous, the annotators proceed to identify incongruous elements, namely those that appear incompatible, and delve into their distinctive attributes or features in both the language and visual modalities. This meticulous examination enables the determination of the concrete target domain and source domain. For instance, in the first photograph of Figure \ref{fig:figure_3}, the advertisement's theme of conserving Earth's resources can be inferred from the image-text combination. By analyzing the image and text, it becomes apparent that there is a domain conflict between the \textit{apple} and the \textit{Earth}. Consequently, the target domain is established as \textit{Earth}, while the source domain is represented by the \textit{apple}. In product advertisements, the theme generally aligns with the promoted product, thus linking the target domain to the product. Likewise, in the second photograph of Figure \ref{fig:figure_3}, the foam of the facial cleanser is metaphorically likened to snow and associated with the miniature skier, resulting in a domain conflict between the two. Therefore, the target domain is identified as \textit{foam}, whereas the source domain pertains to \textit{snow}. 

\subsection{ Annotation Process and Quality Control}

We utilize an approach driven by experts to annotate the data, specifically identifying target and source domains. The annotation team comprises five research graduate students specializing in computational linguistics and possessing expertise in metaphor research. These annotators are organized into three groups: two groups consisting of two members each, and one group consisting of a single member. In cases where the two-member groups fail to reach a consensus, the single-member group participates in the final decision-making process. Completion of the annotation task is considered when there are no disagreements among the group members. However, if discrepancies arise, the single-member group reevaluates the data. If there are differences in annotations across all groups, a collective discussion is conducted to ensure consensus and maintain accuracy and consistency in the annotations. 

In order to improve the quality of the annotations, we implemented several effective measures. We set up stringent standards and documentation for each annotation option, including detailed explanations, abundant examples, and additional notes. Additionally, prior to each annotation session, we conducted training sessions to offer guidance. Throughout the pre-annotation process, we constantly refined the training materials and guidance documents to address any concerns, ensuring that the annotation guidelines were comprehensive and clearly defined before undertaking extensive annotation 
efforts \footnote{Our annotation guideline is on \url{https://gitfront.io/r/GiveATry/nNCeJacwmNpG/CM3D/}. }.

To evaluate the quality of domain labeling, we assessed the agreement of human evaluators with the pre-labeled image-text pairs. Specifically, we randomly selected 200 image-text pairs from the dataset for evaluation. Each domain was evaluated by two annotators, who assigned a value of $1$ if the domain labeling was correct and $0$ otherwise. The human annotators achieved a high level of agreement, with an accuracy of 86\% for the target domain (Cohen's $\kappa$  = 82.35) \cite{cohen1960coefficient} and an accuracy of 84\%  for the source domain (Cohen's $\kappa$ = 80.27), demonstrating the high quality of the dataset.

\subsection{Data Analysis}

Our data is derived from Chinese advertisements, which often use a combination of text, color, graphics, and other elements to convey information about goods and services, ensuring that the public can easily receive and understand the messages. Based on Zhang et al. \cite{zhang-etal-2023-multicmet}, target and source domain items of metaphorical image-text pairs have been categorized into 13 distinct categories.  The source domains show a more dispersed distribution, while the target domains tend to cluster around more concentrated themes and situations.

To analyze the distribution of source and target domains of metaphors in our dataset, we extracted the target and source domains from 6,108 Chinese samples. We used pre-trained GloVe \cite{pennington2014glove} word embeddings to convert the target and source domain words into 96-dimensional vectors. We then applied Uniform Manifold Approximation and Projection (UMAP) \cite{mcinnes2018umap} to reduce the dimensionality of these embeddings and visualize the distribution of the results.

\begin{figure}[b]
    \centering
    \hspace{1cm} 
    \begin{subfigure}[b]{0.45\linewidth}
        \centering
        \includegraphics[height=5cm]{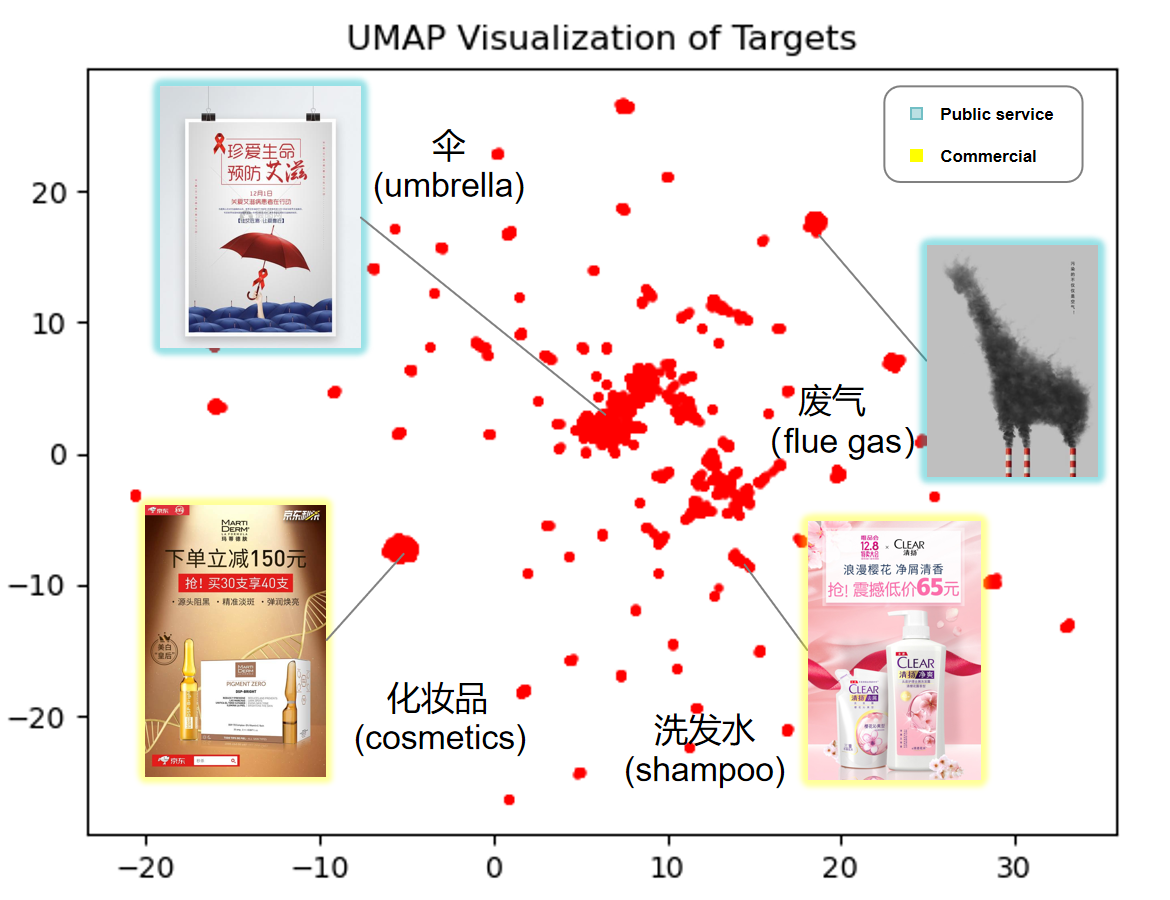}
        \caption{}
        \label{fig:figure_11}
    \end{subfigure}
    \begin{subfigure}[b]{0.45\linewidth}
        \centering
        \includegraphics[height=5cm]{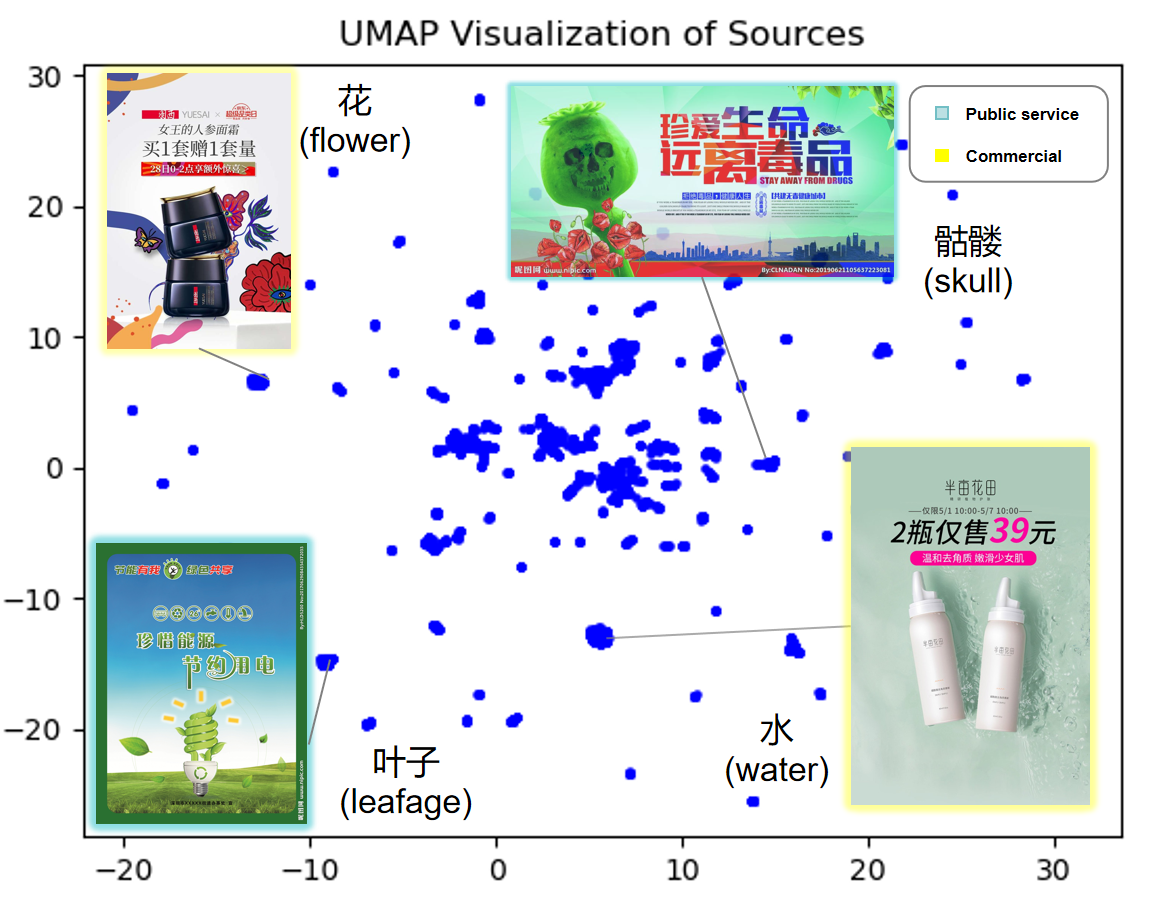}
        \caption{}
        \label{fig:figure_12}
    \end{subfigure}
    \caption{UMAP Visualization of Targets (a) and Sources (b).}
\end{figure}

The figures reveal a significant difference in the distribution of target and source domains between commercial advertisements and public service advertisements, highlighting the varying information and emotional tones conveyed by these two types of ads. Figure \ref{fig:figure_11} shows a relatively concentrated distribution from the perspective of the target domain. Commercial advertisements often focus on promoting and publicizing manufacturers' products, emphasizing their functions or benefits. For instance, facial masks are described as being smooth and moist like water, which means they are often associated with specific products that carry a positive emotional tone, such as cosmetics and facial masks.

On the other hand, public service advertisements are designed to advocate for or prevent certain behaviors or issues. Examples include poppies with skeletons to signify the dangers of drug use, or green playgrounds made of tree trunks and leaves to promote energy conservation and emission reduction. These ads typically convey negative or neutral emotions and, while their distribution is more dispersed compared to commercial ads, they do feature some high-frequency target domain items such as skulls, cars, and trees.

Figure \ref{fig:figure_12} illustrates the distribution from the perspective of the source domain, which appears more dispersed overall. This dispersion is based on product characteristics, such as the moisturizing properties of facial masks or the fragrance of shampoo, as well as advocacy themes, such as the harmful effects of smoking on lungs or the dangers of drunk driving. Consequently, the source domain items are varied, including elements like water, flowers, lungs, and the earth.

Furthermore, we conducted statistical analyses and identified common "source-target" pairs in the advertising dataset. In commercial advertisements, frequent pairs include \textit{skincare products} and \textit{water}, \textit{skincare products} and \textit{flowers}, and \textit{facial masks} and \textit{water}. These metaphors aim to highlight the positive effects of the advertised products. In public service advertisements, examples of common pairs are \textit{skull} and \textit{death}, \textit{drugs} and \textit{skull}, and \textit{car} and \textit{alcohol}. These metaphors serve to caution people to cherish life and steer clear of threats to personal safety. These common combinations that appear in different types of advertising provide some inspiration for our approach.

\section{Methodology}

\subsection{Task Definition}

To gain a deeper understanding of metaphorical expressions, we categorize Chinese metaphorical expressions with finer granularity based on CMT \cite{lakoff1980metaphors} and introduce a novel task: Metaphor Mapping Identification, which involves extracting source and target domain items from metaphorical image-text pairs. The task is illustrated with Figure \ref{fig:figure_1} and Figure \ref{fig:figure_3}:

\begin{verbatim}
    Text: Earth's resources are not like apples, once they are consumed, they are gone.
    Target: earth
    Source: apple
    Sentence: Microscopic foam, cleanses pores.
    Target: foam
    Source: snow
    Text: Smoking damages your lungs
    Target: < model completion >
    Source: < model completion >
\end{verbatim}

Given a metaphorical image and its corresponding textual description, the model is tasked with identifying and extracting the metaphorical source and target domains. Here provides a set of examples, which includes the image, text, and a list of corresponding source and target domain entries. In this case, referring to the example above, the correct output should identify the target domain as \textit{lung} and the source domain as \textit{cigarette}.

\subsection{CoT Prompting-based Metaphor Mapping Identification Model}

In order to enhance metaphorical explanations, the paper introduces a CoT Prompting-based Metaphor Mapping Identification Model (CPMMIM), aiming to generate the target domain and source domain in image-text pairs. In conceptual metaphors, Lakoff \cite{lakoff1980metaphors} refers to the metaphorical object as the source domain and the metaphorical subject as the target domain, with their interaction being the mapping between the two domains. The process of understanding a metaphor involves transferring relevant features from the source domain to the target domain, facilitating comprehension of the target domain. Metaphorical theory suggests that the target domain is often the concept we want to emphasize or explain, while the source domain aids in this understanding. In advertisements, various source domains are utilized to highlight distinct features of the target domain \cite{chang2013missing}. Moreover, we propose that the BLO problem aligns with our metaphorical mapping identification task, exhibiting several insightful parallels. Under the guidance of the BLO framework and CoT prompting, we can leverage the pre-training knowledge of large-scale models and the domain-specific knowledge of small-scale models to obtain the target domain. Subsequently, by utilizing the knowledge of the target domain and reasoning through CoT, we can derive the source domain and establish the mapping relationship in the image-text pair.

\begin{figure}[h]
    \centering
    \includegraphics[height=5.1cm]{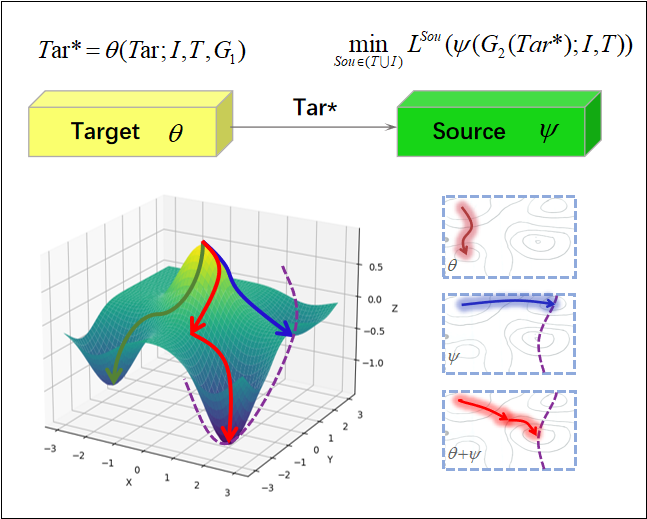}
    \caption{Bi-Level Optimization formulation for metaphor mapping identification.}
    \label{fig:figure_10}
\end{figure}

\subsubsection{Problem Formulation}

Given the dual objectives of source domain identification and target domain identification, and the hierarchical interdependence between these tasks, where source domain identification relies on the accuracy of target domain identification, we have devised a BLO framework grounded in CMT \cite{lakoff1980metaphors} and Stackelberg's theory \cite{liu2022target, ochs2015bilevel} to address the problem. We conceptualize the entire task as a BLO problem. Recognizing that source domain identification is more significant and challenging than target domain identification, and given its reliance on the latter’s accuracy, we designate source domain identification as the Upper-Level problem and target domain identification as the Lower-Level problem. Given an image \(I\) and the corresponding text \(T\), we utilize this BLO framework, formulated as follows:
\begin{equation}
    \min_{\text{Sou} \in \left( T \cup I \right)} \mathcal{L}^{\text{Sou}} \left( \Psi \left( G_2(Tar^*); I, T \right) \right)
\end{equation}

\begin{equation}
    \quad \text{s.t.} \quad Tar^* \in \arg \min_{Tar \in \left( T \cup I \right)} \left( \theta \left( Tar; I, T, G_1 \right) \right)
\end{equation}
where $\mathcal{L}^{\text{Sou}}$ denotes the loss associated with the identification of source domain items, \(I\) represents the image vector derived from the original image, and \(T\) signifies the text vector derived from the original text. \(G_1\) and $G_2$ are the metaphoric features we propose, which are derived from the large model and CoT, respectively. \(\theta\) and \(\Psi\) represent the relevant processing of the MIM module to the fused text and image. Figure \ref{fig:figure_10} illustrates that this bi-level formulation is beneficial for finding the optimal metaphor mapping.

\subsubsection{Model Overview}

\begin{figure*}[t]
    \centering
    \includegraphics[width=0.95\textwidth]{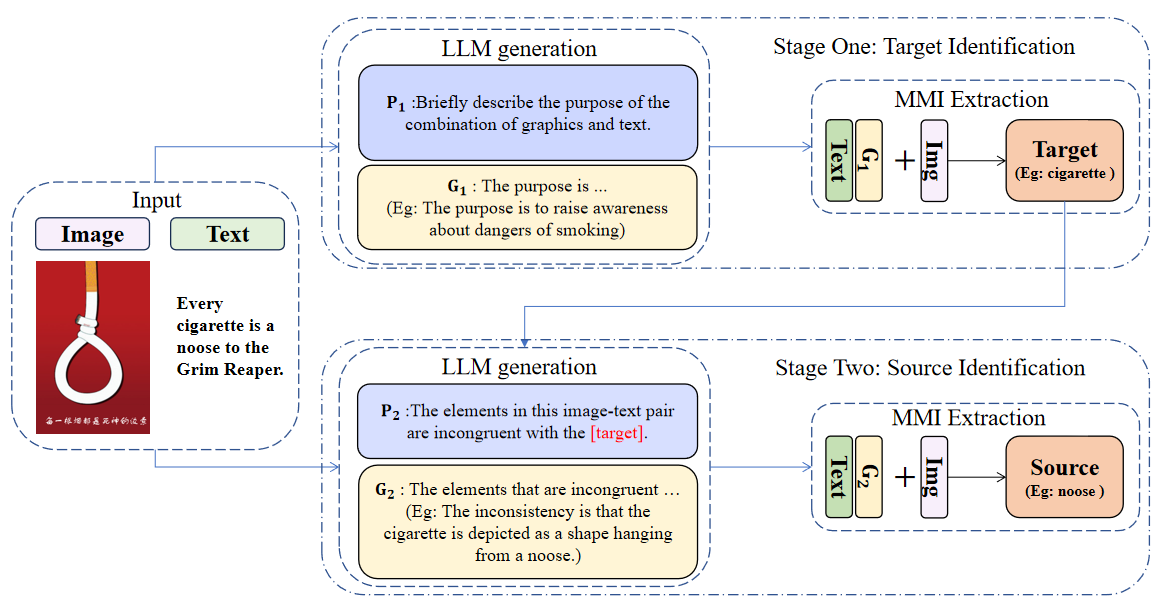}
    \caption{Overview of CPMMIM framework. The entire model operates in two stages, each guided by both the large model and the small model. In this framework, P represents the template prompt for the large model, while G denotes the generated output from the large model. Subsequently, G is incorporated as a new textual feature into the input of the small model.}
    \label{fig:figure_5}
\end{figure*}

The CPMMIM model encompasses two distinct training stages: (i) target domain identification and (ii) source domain identification. While both stages employ the same model architecture, they vary in terms of the input $X$ and output $Y$. The overarching process is depicted in Figure \ref{fig:figure_5}.

The first stage is the target domain identification stage. Considering that the target domain represents the concept to be emphasized or explained in the metaphor, in advertising, the purpose of expressing the product or service is often chosen as the target domain \cite{chang2013missing}. Therefore, when dealing with input images and texts, it is necessary to first inquire about the purpose of the image-text combination theme from the multimodal LLM. This helps ensure that the target domain, which the advertisement seeks to convey, is effectively highlighted, enabling the model to have a more intuitive understanding of the advertisement information. By better integrating the intent and context of the advertisement, the model can infer the accurate target domain.

To achieve this, we input both images and text, and use the following template to ask the multimodal LLM about the theme of the image-text combination:

\noindent
\colorbox{lightgray}{%
    \parbox{\linewidth}{%
$P_1$ = "Briefly describe the purpose of the text-image pair (such as water conservation, promoting shampoo, expressing satire, etc.)"
    }%
}

The multimodal large-scale model generates $G_1$ as output. $G_1$ is then concatenated with the original text $X_{text}$ to form $X^1_{text}$. Finally, the model's input is set to ${X^1}_{input} = \{X^1_{text},\ X_{image}\}$, where $X^1_{text}$ represents the generated text from the first stage and $X_{image}$ represents the original image. The objective of the first stage is to identify the target domain, represented as $T = F(X^1_{input})$, where $T$ corresponds to the identified target domain.

In the source domain identification stage, the $T$ generated in the first stage are appended to a specific prompt, creating the prompt for the second stage, denoted as $P_2$.

The second stage is the source domain identification stage. In metaphor, we use the features of the source domain to aid in understanding the target domain. The source domain and target domain in metaphorical expressions must be two things that, in some sense, have different properties and belong to different domains \cite{lakoff1980metaphors}, which means that the target domain and source domain are two incongruous units. Based on this theoretical foundation, in the source domain identification stage, we attach the $T$ generated in the first stage to specific prompts, thereby creating the prompts for the second stage, represented as $P_2$.

\noindent
\colorbox{lightgray}{%
    \parbox{\linewidth}{%
$P_2$ = "The elements in this image-text pair are incongruent with the [$T$], such as unconventional colors, shapes, and tones."
    }%
}

Then, the generated text, $G_2$, is concatenated with the original text $X_{text}$ to create $X^2_{text}$. Finally, we provide the input ${X^2}_{input} = \{X^2_{text},\ X_{image}\}$ to the source domain identification model to identify the final source domain $S = F(X^2_{input})$.

In the two stages, we train two models with the same architecture independently. We construct the input through a chain of thought prompts to infer the target domain. Based on this, we create new prompts to infer the source domain. Through this step-by-step reasoning process, we ultimately obtain the mapping relationship of metaphorical image-text pairs.

\subsubsection{Metaphor Mapping Identification Module Architecture}

\begin{algorithm}[t]
    \caption{CPMMIM}
    \label{alg:AOA}
    \renewcommand{\algorithmicrequire}{\textbf{Input:}}
    \renewcommand{\algorithmicensure}{\textbf{Output:}}
    \begin{algorithmic}[1]
        \REQUIRE Text input $X_{\text{text}}$, Image input $X_{\text{image}}$
        \ENSURE Identified Target $T$, Source $S$
        
        \STATE \textbf{procedure} $F(X)$
        \STATE \quad $X_{\text{enc}} \leftarrow \text{Encode}(X_{\text{text}}, X_{\text{image}})$
        \STATE \quad $F_{\text{attn}} \leftarrow \text{Attention}(X_{\text{enc}})$
        \STATE \quad $F_{\text{fuse}} \leftarrow \text{Fuse}(X, F_{\text{attn}})$
        \STATE \quad $Y \leftarrow \text{Decoder}(F_{\text{fuse}})$
        \STATE \quad \textbf{return} $Y$
        \STATE \textbf{end procedure}
        
        \STATE $X \leftarrow \{X_{\text{text}}, X_{\text{image}}\}$
        \STATE $G_1 \leftarrow \text{LLMs}(X, P_1)$
        \STATE $X_{\text{text}}^1 \leftarrow X_{\text{text}} \oplus G_1$
        \STATE $X_{\text{input}}^1 \leftarrow \{X_{\text{text}}^1, X_{\text{image}}\}$
        \STATE $T \leftarrow F(X_{\text{input}}^1)$
        \STATE $P_2 \leftarrow \text{ConstructPrompt}(T)$
        \STATE $G_2 \leftarrow \text{LLMs}(X, P_2)$
        \STATE $X_{\text{text}}^2 \leftarrow X_{\text{text}} \oplus G_2$
        \STATE $X_{\text{input}}^2 \leftarrow \{X_{\text{text}}^2, X_{\text{image}}\}$
        \STATE $S \leftarrow F(X_{\text{input}}^2)$
    \end{algorithmic}
\end{algorithm}

\begin{figure}[t]
    \centering
    \includegraphics[width=0.85\textwidth]{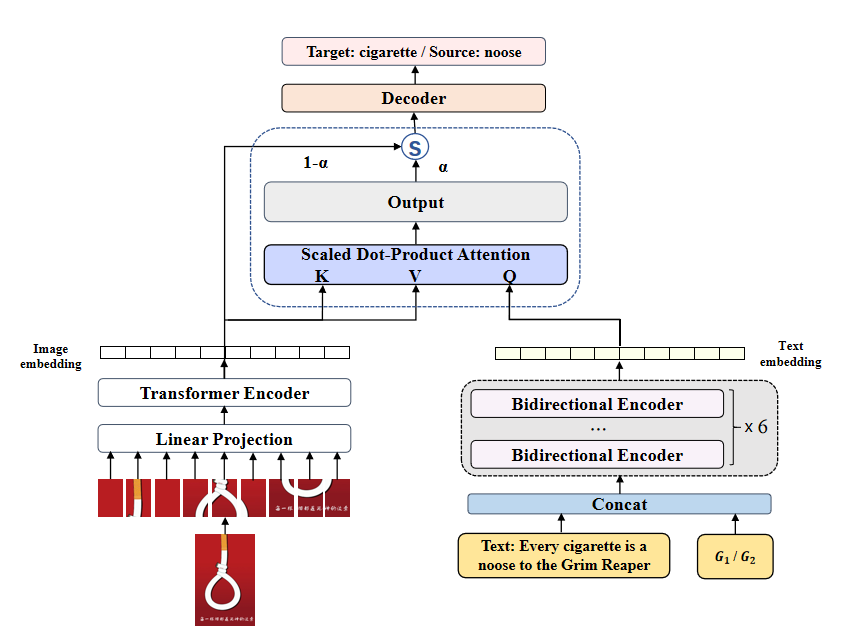}
    \caption{MMI module.}
    \label{fig:figure_6}
\end{figure}%

The Metaphor Mapping Identification (MMI) module consists of three main steps: encoding, interaction, and decoding. This framework is illustrated in Figure \ref{fig:figure_6}, where each part is clearly depicted. Specifically, we input the language text into a Transformer encoder to obtain a representation of the text. Then, we perform interaction and fusion between the language representation and the visual representation to leverage the information from both modalities, where the language text input is $X_{\text{text}} \in \{ X_{\text{text}}^{1}, X_{\text{text}}^{2} \}$ , and the visual input is $X_{\text{image}}$. Finally, the interacted and fused representation is fed into a Transformer decoder for decoding, resulting in the generation of the target text.

\textbf{Encoding}.
The model $F(X)$ simultaneously takes language input and visual input and obtains the text representation $F_{text}$ and image feature $F_{image}$ through the following functions:
\begin{equation}
    \small
    \setlength{\abovedisplayskip}{1pt}
    F_{text} = TextEncoder(X_{text})
\end{equation}

\begin{equation}
    \small
    F_{image} = W_h ImgExtractor(X_{image})
    \setlength{\belowdisplayskip}{1pt}
\end{equation}
\(TextEncoder(·)\) utilizes the hidden states of the last layer in the Bart \cite{lewis-etal-2020-bart} encoder as the language representation, where $F_{\text{text}} \in \mathbb{R}^{n \times d} $, with $n$ denoting the length of the text input and $d $ being the hidden dimension. The working principle of the model combining Bidirectional and Auto-Regressive
Transformers (BART) is to input text into multiple bidirectional Transformer encoders to obtain more context-aware word representations.

\(ImgExtractor(·)\) employs a visual extraction model to obtain patch-level features from the image. The workflow of the Vision Transformer (ViT) model \cite{dosovitskiy2020vit} includes image segmentation, patch flattening, linear mapping, position encoding, input to Transformer encoder, as well as pre-training and fine-tuning steps. After training the ViT model, it can segment the input image data into small patches and flatten them into a one-dimensional sequence of vectors. Through steps like linear mapping, position encoding, and Transformer encoding, it learns to obtain the feature representation of the image. After obtaining the patch-level visual features, we apply a learnable projection matrix $W_h$ to reshape the $ImgExtractor(X_{image})$ into the shape of $X_{text}$, resulting in $F_{image}$. $F_{\text{image}} \in \mathbb{R}^{m \times d} $, where $m$ is the number of patches.

\textbf{Interaction}.
Once the language and image representations have been acquired, we employ a single-head attention network to establish correlations between text tokens and image patches. The query ($Q$), key ($K$), and value ($V$) are derived from $F_{text}$, $F_{image}$, and $F_{image}$, respectively. The attention output $F_{\text{attn}} \in \mathbb{R}^{n \times d} $ is defined as follows:
\begin{equation}
    \small
    \setlength{\abovedisplayskip}{1pt}
    \begin{split}
        F_{attn} = softmax\left(\frac{QK^T}{\sqrt{d_k}}\right)V
    \end{split}
    \setlength{\belowdisplayskip}{1pt}
\end{equation}
Subsequently, we incorporate the gated fusion mechanism to fuse $F_{text}$ and $F_{image}$. The gate mechanism can filter out noise information in both images and text. The fused output, $F_{\text{fuse}} \in \mathbb{R}^{n \times d} $, is obtained via:
\begin{equation}
    \small
    \setlength{\abovedisplayskip}{1pt}
    \begin{split}
        \alpha = Sigmoid(W_tF_{text} + W_vF_{attn})
    \end{split}
    \setlength{\belowdisplayskip}{1pt}
\end{equation}
\begin{equation}
    \small
    \setlength{\abovedisplayskip}{1pt}
    \begin{split}
        F_{fuse} = (1-\alpha) · F_{text} + \alpha·F_{attn}
    \end{split}
    \setlength{\belowdisplayskip}{1pt}
\end{equation}
where $W_t$ and $W_v$ are learnable parameters.

\textbf{Decoding}.  
In a decoder based on the Transformer architecture, the fused representation $F_{fuse}$ is passed as input to multiple autoregressive Transformer decoders for generating text output word by word. This approach allows the BART model to effectively utilize contextual information for understanding the text while ensuring syntactically and semantically correct generated text. Each decoder layer consists of a self-attention mechanism and feed-forward neural networks. The self-attention mechanism enables the model to attend to other positions in the input sequence with weighted attention during each prediction, capturing global contextual information. With the collaborative work of multiple decoder layers, the Transformer decoder can efficiently transform the fused representation $F_{fuse}$ into predicted results in the target domain corresponding to the source domain.

\section{Performance Evaluation}

\subsection {Research Questions}
Based on the gaps identified in previous studies, this paper aims to address the following research questions:

\begin{itemize}
    \item \textbf{RQ1}: How does the performance of large-scale language models compare in the task of metaphor mapping identification?
\item \textbf{RQ2}: How do unimodal models (text-only or image-only) perform compared to multimodal models in metaphor mapping identification?
\item \textbf{RQ3}: Is the extraction of the source domain more challenging than the extraction of the target domain in metaphor mapping identification?
\item \textbf{RQ4}: Does the inclusion of prompts in both stages (target and source domain identification) improve the model's performance?
    
\end{itemize}

To address these research questions, we have developed a comprehensive methodology centered around our proposed CoT Prompting-based Metaphor Mapping Identification Model (CPMMIM). Our approach involves designing experiments that evaluate the performance of various models—including large-scale language models and both unimodal and multimodal models—on the task of metaphor mapping identification. We also perform ablation studies to assess the impact of including prompts in both stages of our model.

The following sections provide detailed descriptions of our task definition, model architecture, and experimental setup. This comprehensive methodology sets the foundation for the performance evaluation presented in Section 5, where we discuss the results in the context of our research questions

\subsection{Experimental Settings}

We employed the PyTorch \cite{paszke2019pytorch} framework and utilized the encoder-decoder architecture of the bart-large-chinese \cite{shao2021cpt} model. For image feature extraction, we employed the ViT-L/16 model \cite{dosovitskiy2020vit}. Text processing was performed using the bert-base-chinese \cite{devlin-etal-2019-bert} tokenizer and encoder to handle Chinese text. To expedite model training and inference, we leveraged the GeForce RTX 4090 GPU. During the training process, the cross-entropy loss function was utilized as the objective function to minimize the discrepancy between the predicted results and the ground truth labels. We incorporated prompts into the Qwen-vl \cite{Qwen-VL} model to acquire knowledge. Due to the lack of model details and API call limitations for some large models such as GPT-4 \footnote{\url{https://platform.openai.com/docs/models/gpt-4-and-gpt-4-turbo}} and Qwen-vl-plus \cite{Qwen-VL}, they are not suitable for handling large-scale datasets. Therefore, we did not use these models for training, but instead used them as baselines for evaluation on the test set. In addition to these settings, we made further adjustments according to the specific experimental requirements, which are detailed in Table \ref{tab:table_2}.

\begin{table}[!ht]
    \small
    \centering
    \caption{Hyperparameters.}
    \begin{tabular}{ccccccc}
    \toprule[1pt]
        \textbf{Hyper-parameter} & Dropout & Input length & Output length & Epoch & Learning rate & Batch size \\ 
    \midrule[0.5pt]
        \textbf{Value} & 0.3 & 32 & 16 & 8 & 5e-6 & 16 \\ 
    \bottomrule[1pt]
    \end{tabular}
    
    \label{tab:table_2}
\end{table}

\subsection{Baselines}

To comprehend metaphors, we propose two tasks: multimodal target domain generation and multimodal source domain generation. 
In line with the work of \cite{zhang-etal-2021-multimet}, our experiments covered three modalities: text, image, and a combination of both. The inference process of all large-scale models is based on few-shot prompting, which concatenates two in-context examples from the training set with the same context before the test instance.

\textbf{BART} \cite{lewis-etal-2020-bart}: BART is a pre-trained sequence-to-sequence model. In our experiments, we used bart-large-chinese \cite{lewis-etal-2020-bart} as the baseline model.

\textbf{GPT-3.5-Turbo (GPT-3.5)} \footnote{\url{https://platform.openai.com/docs/models/gpt-3-5}}: We use GPT-3.5-turbo for inference, which has been pretrained on large-scale diverse datasets, enabling it to possess extensive knowledge and language understanding capabilities.

\textbf{ERNIE-Bot} \cite{cheng2023evaluating}: ERNIE-Bot is a new generation of Chinese language model for knowledge enhancement, and we access its 3.5 version through its online open API. By inputting the text from image-text pairs into the model and providing prompts to guide its output in the target and source domains.

\textbf{LLaVA} \cite{liu2024visual}: LLaVA is a large-scale multimodal model (LMM) developed by researchers by connecting the open-source visual encoder CLIP with the language decoder LLaMA. In this paper, we selected LLaVA-v1.5-7b as the baseline model. By inputting images alone or in combination with text, and providing specific prompts to guide the model to generate output in the target and source domains.

\textbf{Qwen-VL-plus and Qwen-VL} \cite{Qwen-VL}: Qwen-VL and Qwen-VL-plus are large-scale visual language models that support multiple languages, including Chinese and English. They significantly enhance the ability to process text in images. Qwen-VL-Plus has undergone significant upgrades in detail recognition and text recognition capabilities, supporting ultra-high pixel resolution and images with arbitrary aspect ratios. We input data through two methods: using only images or using both images and text.

\textbf{GPT-4-Vision-preview (GPT-4)}: GPT-4 is an exceptional multimodal large-scale model with 1 trillion parameters, enabling advanced image comprehension and reasoning capabilities. As the model has not yet been made publicly available, we utilize it through API calls. We employ two methods of data input: solely using images or using a combination of both images and text.

\textbf{BART+ViT} \cite{dosovitskiy2020vit}: BART+ViT combines the BART language model with the ViT visual encoder, enabling it to handle both text and images simultaneously. During the training phase, we directly concatenate the image-text pairs as input to the model and fine-tune it using the training dataset. We input data through two methods: using only images or using both images and text. In the inference phase, we directly output the results in the target and source domains.

\textbf{BART+DETR} \cite{carion2020end}: BART+DETR has a similar training process to BAET+ViT, with the difference being that the visual encoder is replaced by DETR.

\subsection{Evaluation Metrics}

The task of generating domain-specific text differs from conventional text generation tasks such as translation or summarization, as the generated text for domain-specific tasks is typically much shorter, usually consisting of only 2-5 tokens. As a result, traditional n-gram-based evaluation metrics, such as BLEU and ROUGE, may not be applicable, as these metrics typically consider longer text segments.
We use three evaluation metrics to assess the quality of the model's generated results: BertScore \cite{bert-score}, accuracy, and human evaluation. These three metrics can evaluate the quality of the results from different perspectives, considering relative accuracy, absolute accuracy, and overall quality, respectively.

\textbf{BertScore} \cite{bert-score}. It is based on the BERT pre-trained model and utilizes contextual embeddings to represent sentences. It computes the cosine similarity between two sentences and captures semantic similarity that traditional n-gram methods may overlook. In this paper, we primarily utilize the F1 Score of BertScore to evaluate the relative accuracy of the generated domains. F1 Score balances precision and recall, providing a single performance metric. 

\textbf{Accuracy.}
Only when the model's generated domain exactly matches the annotated domain is it considered completely correct; otherwise, it is deemed incorrect. This evaluation metric aids in assessing the model's ability to generate domains with a high degree of accuracy.

\textbf{Human-Evaluation.}
Due to the variations in expression and levels of precision across different domains, it is challenging to obtain accurate scores directly from automated evaluations. Therefore, to ensure a more reliable evaluation of the model's generated outputs, we conducted manual checks. In challenging cases, annotators followed the Metaphor Identification Procedure (MIP) \cite{steen2010method}, analyzing the underlying, more fundamental meanings of words and determining if these meanings aligned with the target and source domains as predicted \cite{wachowiak-gromann-2023-gpt}. If the annotator judges that the model's inferred target domain matches the annotated target domain for the given image-text pair, it is considered a correct prediction by the model for that pair's target domain; otherwise, it is considered a prediction error. Accuracy is calculated by determining the percentage of correct predictions in the target domain relative to the total dataset, yielding the H-E score of the model in the target domain. A similar evaluation methodology is applied to the source domain. Two annotators (who are experts in computational linguistics and metaphor research) independently assessed the model's outputs and discussed any discrepancies with a third annotator (one of the authors). This manual evaluation was implemented to ensure a comprehensive and accurate assessment of the generated domains. During this process, we utilized kappa values to evaluate the agreement among annotators. Higher kappa values indicate better agreement between annotators.

\begin{table*}[b]
  \caption{Results of target domain in CM3D.}
\begin{tabular}{cccccccccc}
\toprule[1pt]
                            &                    & Acc   & BertScore & H-E   & Cohen’s Kappa         \\ \hline
\multirow{3}{*}{Text}       & BART                & 13.15 & 70.63     & 40.49 & 76.29(Substantial)    \\
                            & GPT-3.5             & 8.69  & 64.95     & 33.78 & 87.29(Almost perfect) \\
                            & ERNIE-Bot           & 4.26  & 30.15     & 23.11 & 86.27(Almost perfect) \\ \hline
\multirow{6}{*}{Image}      & GPT-4                & 18.20 & 69.82     & 38.36 & 78.15(Substantial)    \\
                            & LLaVA               & 5.25  & 55.49     & 27.38 & 87.56(Almost perfect) \\
                            & Qwen-VT             & 9.84  & 56.94     & 35.74 & 85.46(Almost perfect) \\
                            & Qwen-VT-plus        & 9.34  & 57.94     & 37.70 & 84.68(Almost perfect) \\
                            & Bart+DETR(w/o text) & 16.07 & 76.75     & 45.90 & 74.73(Substantial)    \\
                            & Bart+ViT(w/o text)  & 27.05 & 78.35     & 48.36 & 69.54(Substantial)    \\ \hline
\multirow{7}{*}{Text+Image} & GPT-4                & 20.16 & 71.78     & 40.50 & 75.81(Substantial)     \\
                            & LLaVA               & 8.85  & 66.37     & 36.89 & 88.48(Almost perfect) \\
                            & Qwen-VT             & 10.66  & 46.77     & 37.21 & 87.56(Almost perfect) \\
                            & Qwen-VT-plus        & 11.97  & 66.94     & 38.52 & 85.46(Almost perfect) \\
                            & Bart+DETR           & 28.52 & 76.54     & 50.09 & 74.83(Substantial)     \\
                            & Bart+ViT            & 32.13 & 80.34     & 53.28 & 68.56(Substantial)    \\
                            & CPMMIM               & \pmb{36.23} & \pmb{84.87}     & \pmb{58.52} & 71.68(Substantial)     \\ 
    \bottomrule[1pt]
    \end{tabular}
  
    \label{tab:table_3}
\end{table*}

\begin{table*}[]
\caption{Results of source domain in CM3D.}
\begin{tabular}{cccccccccc}
\toprule[1pt]
                    
                            &                     & Acc   & BertScore & H-E   & Cohen’s Kappa         \\ \hline
\multirow{3}{*}{Text}       & BART                 & 10.33 & 68.26     & 32.45 & 80.38(Almost perfect) \\
                            & GPT-3.5             & 7.37 & 50.78     & 30.16 & 89.26(Almost perfect) \\
                            & ERNIE-Bot           & 5.41  & 39.98     & 25.25 & 87.29(Almost perfect) \\ \hline
\multirow{6}{*}{Image}      & GPT-4               & 16.48 & 65.29     & 33.39 & 79.29(Substantial)    \\
                            & LLaVA               & 4.92  & 63.85     & 21.80 & 92.67(Almost perfect) \\
                            & Qwen-VT             & 4.59  & 55.76     & 28.52 & 88.46(Almost perfect) \\
                            & Qwen-VT-plus        & 5.73  & 57.16     & 29.84 & 81.26(Almost perfect) \\
                            & Bart+DETR(w/o text) & 12.31 & 74.44     & 27.87 & 78.18(Substantial)    \\
                            & Bart+ViT(w/o text) & 17.87 & 76.08     & 32.58 & 75.37(Substantial)    \\ \hline
\multirow{7}{*}{Text+Image} & GPT-4                & 16.72 & 66.62     & 31.64 & 76.19(Substantial)    \\
                            & LLaVA               & 5.56  & 58.83     & 20.46 & 87.68(Almost perfect) \\
                            & Qwen-VT             & 3.95  & 55.27     & 28.20 & 92.67(Almost perfect) \\
                            & Qwen-VT-plus        & 5.86  & 59.76     & 30.98 & 88.46(Almost perfect) \\
                            & Bart+DETR           & 24.59 & 75.15     & 44.43 & 77.26(Substantial)    \\
                            & Bart+ViT            & 25.38 & 76.90     & 46.89 & 74.92(Substantial)    \\
                            & CPMMIM               & \pmb{28.47} & \pmb{80.84}     & \pmb{50.49} & 75.43(Substantial)    \\ 
    \bottomrule[1pt]
    \end{tabular}
    
    \label{tab:table_9}
\end{table*}

\subsection{Results and Discussion}

Table \ref{tab:table_3} and Table \ref{tab:table_9} present the main experimental results, which show that our model outperforms other baseline methods. These experimental results validate the effectiveness of multimodality and demonstrate the effectiveness of using prompt-based CoT for understanding metaphors. Table \ref{tab:table_4} shows the results of ablation experiments, which confirm the effectiveness of using prompts to guide the model to generate target and source domain related information in accordance with human thinking patterns. In addition, we conducted a more detailed analysis of the generated results of models for target and source domains. We present the experimental results and discuss them in relation to our research questions below. 

\textbf{RQ1: How does the performance of large models?}

\begin{figure*}[t]
    \centering
    \includegraphics[width=0.9\textwidth]{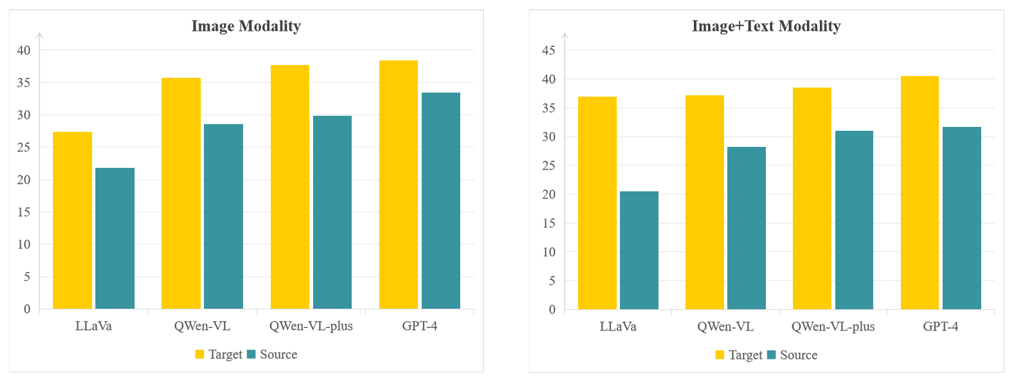}
    \caption{Results of multimodal large models in target and source domains.}
    \label{fig:figure_7}
\end{figure*}

The data in the table shows that the performance of directly generating target and source domains using large models or performing few-shot learning with large models is much worse than using smaller models such as BART. Through comparing various large models, we found that GPT-3.5 performs better in text-only modality. In contrast, as seen from Figure \ref{fig:figure_7}, the GPT-4 model performs the best for image-text and multimodal tasks, which may be related to the size of the model's parameters. The larger the number of parameters, the better the model performs on this task. Additionally, it is observed that the presence of a text modality has minimal impact on the results, as the large model itself possesses certain capabilities for extracting text from images.

We also conducted a detailed analysis of the errors made by the large models, and found three main phenomena in the domain generation task. The first phenomenon is that due to the complexity of metaphor recognition tasks, the large model mistakenly recognizes metaphorical image-text pairs as non-metaphorical, resulting in the failure to generate target and source domains. The second phenomenon is that due to the complexity of the target and source domain generation task, the model may misunderstand and confuse the concepts of the two domains, thus making mistakes by treating the source domain as the target domain. The third phenomenon is that the model may produce hallucinations during the analysis process, incorporating redundant knowledge that leads to generated target and source domains unrelated to the image-text pairs.

\textbf{RQ2: How do unimodal models perform compared to multimodal models?} 

Based on the results, it is evident that multimodal models perform much better than unimodal models. For large models, the performance of the text-only modality is relatively poor. Through specific case analysis, it was found that the model generates strong hallucination when lacking image features, resulting in nonsensical background information based on the text. Multimodal models with image modality and multimodal modality have similar performances, as the text in the dataset is extracted from the images, and the multimodal models have developed excellent OCR capabilities to analyze the text features in conjunction with the image features.

By comparing the small models of the text-only modality and the image-only modality, it can be seen that the generated results of the multimodal models are optimal. This indicates that both image and text features are important during small model training. The use of the Bart+ViT multimodal model improves the manually evaluated scores in the target domain and source domain by 12.79 and 14.44 respectively, as compared to just using the Bart model.

\textbf{RQ3: Is the extraction of the source domain more difficult than the extraction of the target domain?}

Based on the results, it can be observed that the target domain usually outperforms the source domain, indicating that extracting information from the image and text pairs in the source domain is more challenging. Through specific examples, it can be concluded that the purpose of advertising is typically to convey specific information or to prompt audiences to take certain actions. In advertisements, in order to deliver information more quickly, these intentions and purposes are often directly reflected in prominent objects or text in the images. The target domain is closely related to the purpose of the advertisement, for example, in a product advertisement, the target domain is likely to be related to the product being promoted. On the other hand, the source domain in advertisements has a wide range of options, which are adjusted based on the characteristics of the target domain that the advertisement aims to highlight, and some source domains are also related to cultural background, which requires a deeper understanding of the context of the image and text to determine.

\begin{table*}[]
\caption{Ablation results of CPMMIM.}
\setlength\tabcolsep{3pt}
\begin{tabular}{ccccccccc}
\toprule[1pt]
          & \multicolumn{4}{c}{Target}                     & \multicolumn{4}{c}{Source}                        \\ \cline{2-9} 
          & Acc   & BertScore & H-E   & Cohen’s Kappa      & Acc   & BertScore & H-E   & Cohen’s Kappa         \\ \hline
CPMMIM     & 36.23 & 84.87     & 58.52 & 71.68(Substantial) & 28.47 & 80.84     & 50.49 & 75.43(Substantial)    \\
w/o $PG_1$ & 32.15 & 80.34     & 53.24 & 68.56(Substantial) & 26.84 & 77.78     & 47.70 & 80.53(Almost perfect) \\
w/o $PG_2$ & 36.23 & 84.87     & 58.52 & 71.68(Substantial) & 26.59 & 78.38     & 48.52 & 81.24(Almost perfect) \\ 
    \bottomrule[1pt]
    \end{tabular}
    
    \label{tab:table_4}
\end{table*}

\textbf{RQ4: Is the prompt for both stages effective?}

To validate the effectiveness of incorporating relevant knowledge generated by LLMs into both the target and source domains, we conducted ablation experiments. In these experiments, we removed $PG_1$ and $PG_2$ separately in the two stages and observed the model's generation results. 
According to the results in Table \ref{tab:table_4}, it can be observed that the recognition performance is best for both the target and source domains with the  prompt. This indicates that both $PG_1$ and $PG_2$ stages make significant contributions to metaphor comprehension. This indicates that relying solely on a single text generation model, such as BART, and an image feature extractor, such as ViT, is not accurate enough for understanding metaphorical language production. To improve the recognition accuracy of both target and source domains, it is necessary to combine the background knowledge and deep understanding of images provided by LLMs with specific metaphorical knowledge learned from small-scale models trained on the training set.

The table shows that removing the $PG_1$ stage not only affects the recognition accuracy of the target domain but also has an impact on the source domain. This suggests that there is a mutual influence between the target and source domains, and improving the recognition accuracy of the target domain can enhance the recognition accuracy of the source domain. The removal of the $PG_2$ stage only affects the recognition accuracy of the source domain, leading to a decrease of 1.9\% in the H-E of the source domain.

\subsection{Error Analysis}

To gain a deeper insight into the performance of our model and to support future research efforts, we conducted a manual examination of randomly selected instances where our method generated incorrect responses. We scrutinized 50 samples with errors and identified the common mistakes as misconceptions and cultural misinterpretations.

The first type of error occurs when certain domains cannot be directly inferred from the image or text alone and require contextual understanding that combines common knowledge and the image-text pairing. However, due to the model's insufficient understanding of images, it may mistakenly extract the domain incorrectly. In the given example image, the purpose is to promote the value of cherishing food. In the image, a farmer is depicted bending over to insert rice into a bowl of food. According to common knowledge, the rice should be planted in soil. Therefore, the image metaphorically compares the rice to soil, emphasizing the importance of conserving food. The target domain is \textit{food} and the source domain is \textit{soil}. Upon inputting this image-text pair, the large model provides $G_1$ as "The purpose of this image is to remind people to cherish food and oppose waste." The model correctly identifies the target domain as \textit{food}. However,$G_2$ generated by the large model states, "By incorporating the scene of farmers cultivating into each grain of food, the image vividly showcases the process of food production, highlighting the arduousness of food." The final source domain identified by the model is \textit{farmer}. It is evident that the model simply understands the image as depicting the scene of farmers cultivating merged with each grain of food, without grasping the conflict that rice should be planted in soil, not on top of food. This leads to an incorrect identification of the source domain.

The second type of error involves the understanding of metaphors, which is closely related to cultural background. There are significant differences in how different cultures interpret symbols and signs, and a model's lack of understanding of specific cultural backgrounds can lead to missing key elements in the analysis of image-text information, thereby affecting the accuracy of domain extraction. For example, in Chinese culture, the tripod (Ding) symbolizes honesty and the importance of keeping promises. In this image, to emphasize the core value of honesty, the designer placed a golden tripod on a triangular base. In this case, the target domain of the image-text pair should be identified as \textit{honesty}, with the source domain being the \textit{tripod}. However, when this image-text pair is analyzed by a large model, the feedback obtained from the model, $G_1$, states, "The main purpose of this image is to convey the value of honesty." At this point, the model correctly identifies the target domain as \textit{honesty}. But the feedback obtained from the model, $G_2$, states: "honesty is highlighted in red calligraphy." This indicates that the model mistook \textit{calligraphy} for the source domain, clearly overlooking the critical symbol of the \textit{tripod} and its connection to \textit{honesty}. This error reflects the model's deficiency in domain recognition.

 \begin{figure}[H]
	 	\centering  
	 	\includegraphics[width=0.75\textwidth] {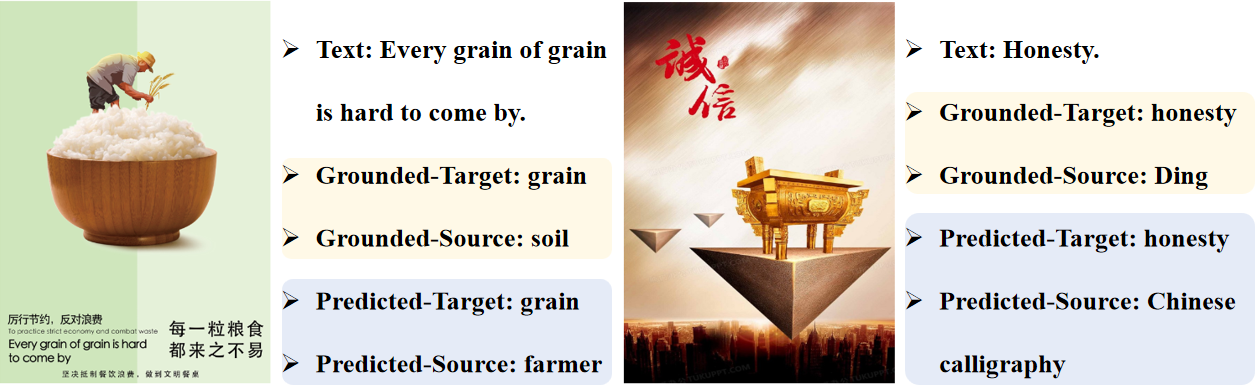}    
	 	\caption{Two examples of identification errors.}
	 	\label{fig:figure_8}
	 \end{figure}

\section{Conclusion}

This paper introduces the Chinese Multimodal Metaphor Mapping Dataset (CM3D), a valuable resource featuring detailed annotations of specific target and source domains in the Chinese language. By providing this dataset, we aim to facilitate further research on metaphors, particularly in non-English linguistic contexts. Inspired by the emerging concept of CoT and Bi-Level Optimization (BLO), we propose the CoT Prompting-based Metaphor Mapping Identification Model (CPMMIM), which simulates the complex human thinking process involved in generating target and source domains for metaphors.Our experimental results validate the effectiveness of CPMMIM, demonstrating its capability to identify and map target and source domains accurately. To foster collaboration and progress in the field, we have made both the code and the CM3D dataset publicly available\footnote{\url{https://gitfront.io/r/GiveATry/nNCeJacwmNpG/C3MD/}}. This openness encourages researchers to explore and contribute to the evolving realm of multimodal metaphor understanding, not only in Chinese but also in diverse linguistic and cultural contexts.

While our work makes significant contributions, there are limitations that future research could address. For instance, expanding the dataset to include a wider variety of metaphors and testing the model across different languages and cultures could enhance the generalizability of our findings. In essence, this paper provides a comprehensive framework, dataset, and model, offering valuable insights into the study of metaphors in natural language processing. We hope that our work will inspire further research and development in this area, advancing the understanding of metaphoric language in natural language processing.

\bibliographystyle{unsrt}
\bibliography{sample-base}

\begin{thebibliography}{10}

\bibitem{steen2010method}
Gerard Steen, Lettie Dorst, Berenike Herrmann, Anna Kaal, Tina Krennmayr, and Trijntje Pasma.
\newblock A method for linguistic metaphor identification from mip to mipvu preface.
\newblock {\em Method for linguistic metaphor identification: from MIP To MIPVU}, 14:9--20, 2010.

\bibitem{lakoff1980metaphors}
G.~Lakoff and M.~Johnson.
\newblock {\em Metaphors We Live By}.
\newblock University of Chicago Press, 1980.

\bibitem{xubomm2024}
Bo~Xu, Junzhe Zheng, Jiayuan He, Yuxuan Sun, Hongfei Lin, Liang Zhao, and Feng Xia.
\newblock Generating multimodal metaphorical features for meme understanding.
\newblock In {\em Proceedings of the 32nd ACM International Conference on Multimedia}, MM'24, page 447–455, 2024.

\bibitem{xubosigir22}
Bo~Xu, Tingting Li, Junzhe Zheng, Mehdi Naseriparsa, Zhehuan Zhao, Hongfei Lin, and Feng Xia.
\newblock Met-meme: A multimodal meme dataset rich in metaphors.
\newblock In {\em Proceedings of the 45th International ACM SIGIR Conference on Research and Development in Information Retrieval}, SIGIR'22, page 2887–2899, 2022.

\bibitem{6594800}
Feng Xia, Nana~Yaw Asabere, Ahmedin~Mohammed Ahmed, Jing Li, and Xiangjie Kong.
\newblock Mobile multimedia recommendation in smart communities: A survey.
\newblock {\em IEEE Access}, 1:606--624, 2013.

\bibitem{forceville2009multimodal}
Charles Forceville, Eduardo Urios-Aparisi, et~al.
\newblock {\em Multimodal metaphor}, volume~11.
\newblock Mouton de Gruyter Berlin, 2009.

\bibitem{kuo-carpuat-2020-evaluating}
Kevin Kuo and Marine Carpuat.
\newblock Evaluating a {B}i-{LSTM} model for metaphor detection in {TOEFL} essays.
\newblock In Beata~Beigman Klebanov, Ekaterina Shutova, Patricia Lichtenstein, Smaranda Muresan, Chee Wee, Anna Feldman, and Debanjan Ghosh, editors, {\em Proceedings of the Second Workshop on Figurative Language Processing}, pages 192--196, Online, July 2020. Association for Computational Linguistics.

\bibitem{liu-etal-2020-metaphor}
Jerry Liu, Nathan O{'}Hara, Alexander Rubin, Rachel Draelos, and Cynthia Rudin.
\newblock Metaphor detection using contextual word embeddings from transformers.
\newblock In Beata~Beigman Klebanov, Ekaterina Shutova, Patricia Lichtenstein, Smaranda Muresan, Chee Wee, Anna Feldman, and Debanjan Ghosh, editors, {\em Proceedings of the Second Workshop on Figurative Language Processing}, pages 250--255, Online, July 2020. Association for Computational Linguistics.

\bibitem{choi-etal-2021-melbert}
Minjin Choi, Sunkyung Lee, Eunseong Choi, Heesoo Park, Junhyuk Lee, Dongwon Lee, and Jongwuk Lee.
\newblock {M}el{BERT}: Metaphor detection via contextualized late interaction using metaphorical identification theories.
\newblock In Kristina Toutanova, Anna Rumshisky, Luke Zettlemoyer, Dilek Hakkani-Tur, Iz~Beltagy, Steven Bethard, Ryan Cotterell, Tanmoy Chakraborty, and Yichao Zhou, editors, {\em Proceedings of the 2021 Conference of the North American Chapter of the Association for Computational Linguistics: Human Language Technologies}, pages 1763--1773, Online, June 2021. Association for Computational Linguistics.

\bibitem{Yang2013ContextualEO}
Fan-Pei~Gloria Yang, Kailyn A.~L. Bradley, Madiha Huq, Dai-Lin Wu, and Daniel~C. Krawczyk.
\newblock Contextual effects on conceptual blending in metaphors: An event-related potential study.
\newblock {\em Journal of Neurolinguistics}, 26:312--326, 2013.

\bibitem{mohammad-etal-2016-metaphor}
Saif Mohammad, Ekaterina Shutova, and Peter Turney.
\newblock Metaphor as a medium for emotion: An empirical study.
\newblock In Claire Gardent, Raffaella Bernardi, and Ivan Titov, editors, {\em Proceedings of the Fifth Joint Conference on Lexical and Computational Semantics}, pages 23--33, Berlin, Germany, August 2016. Association for Computational Linguistics.

\bibitem{rosen2018computationally}
Zachary Rosen.
\newblock Computationally constructed concepts: A machine learning approach to metaphor interpretation using usage-based construction grammatical cues.
\newblock In {\em Proceedings of the Workshop on Figurative Language Processing}, pages 102--109, 2018.

\bibitem{wachowiak-gromann-2023-gpt}
Lennart Wachowiak and Dagmar Gromann.
\newblock Does {GPT}-3 grasp metaphors? identifying metaphor mappings with generative language models.
\newblock In Anna Rogers, Jordan Boyd-Graber, and Naoaki Okazaki, editors, {\em Proceedings of the 61st Annual Meeting of the Association for Computational Linguistics (Volume 1: Long Papers)}, pages 1018--1032, Toronto, Canada, July 2023. Association for Computational Linguistics.

\bibitem{shutova2017multilingual}
Ekaterina Shutova, Lin Sun, Elkin~Dar{\'\i}o Guti{\'e}rrez, Patricia Lichtenstein, and Srini Narayanan.
\newblock Multilingual metaphor processing: Experiments with semi-supervised and unsupervised learning.
\newblock {\em Computational Linguistics}, 43(1):71--123, 2017.

\bibitem{zhang-etal-2021-multimet}
Dongyu Zhang, Minghao Zhang, Heting Zhang, Liang Yang, and Hongfei Lin.
\newblock {M}ulti{MET}: A multimodal dataset for metaphor understanding.
\newblock In Chengqing Zong, Fei Xia, Wenjie Li, and Roberto Navigli, editors, {\em Proceedings of the 59th Annual Meeting of the Association for Computational Linguistics and the 11th International Joint Conference on Natural Language Processing (Volume 1: Long Papers)}, pages 3214--3225, Online, August 2021. Association for Computational Linguistics.

\bibitem{mohler2016introducing}
Michael Mohler, Mary Brunson, Bryan Rink, and Marc Tomlinson.
\newblock Introducing the lcc metaphor datasets.
\newblock In {\em Proceedings of the Tenth International Conference on Language Resources and Evaluation (LREC'16)}, pages 4221--4227, 2016.

\bibitem{dodge2015metanet}
Ellen~K Dodge, Jisup Hong, and Elise Stickles.
\newblock Metanet: Deep semantic automatic metaphor analysis.
\newblock In {\em Proceedings of the Third Workshop on Metaphor in NLP}, pages 40--49, 2015.

\bibitem{mao2022metapro}
Rui Mao, Xiao Li, Mengshi Ge, and Erik Cambria.
\newblock Metapro: A computational metaphor processing model for text pre-processing.
\newblock {\em Information Fusion}, 86:30--43, 2022.

\bibitem{ge2022explainable}
Mengshi Ge, Rui Mao, and Erik Cambria.
\newblock Explainable metaphor identification inspired by conceptual metaphor theory.
\newblock In {\em Proceedings of the AAAI Conference on Artificial Intelligence}, volume~36, pages 10681--10689, 2022.

\bibitem{su2024efficient}
Chang Su, Xingyue Wang, Shupin Liu, and Yijiang Chen.
\newblock Efficient visual metaphor image generation based on metaphor understanding.
\newblock {\em Neural Processing Letters}, 56(3):150, 2024.

\bibitem{leong-etal-2020-report}
Chee Wee~(Ben) Leong, Beata Beigman~Klebanov, Chris Hamill, Egon Stemle, Rutuja Ubale, and Xianyang Chen.
\newblock A report on the 2020 {VUA} and {TOEFL} metaphor detection shared task.
\newblock In Beata~Beigman Klebanov, Ekaterina Shutova, Patricia Lichtenstein, Smaranda Muresan, Chee Wee, Anna Feldman, and Debanjan Ghosh, editors, {\em Proceedings of the Second Workshop on Figurative Language Processing}, pages 18--29, Online, July 2020. Association for Computational Linguistics.

\bibitem{10.1145/3551349.3559555}
Toufique Ahmed and Premkumar Devanbu.
\newblock Few-shot training llms for project-specific code-summarization.
\newblock In {\em Proceedings of the 37th IEEE/ACM International Conference on Automated Software Engineering}, ASE '22, New York, NY, USA, 2023. Association for Computing Machinery.

\bibitem{paranjape-etal-2021-prompting}
Bhargavi Paranjape, Julian Michael, Marjan Ghazvininejad, Hannaneh Hajishirzi, and Luke Zettlemoyer.
\newblock Prompting contrastive explanations for commonsense reasoning tasks.
\newblock In Chengqing Zong, Fei Xia, Wenjie Li, and Roberto Navigli, editors, {\em Findings of the Association for Computational Linguistics: ACL-IJCNLP 2021}, pages 4179--4192, Online, August 2021. Association for Computational Linguistics.

\bibitem{liu-etal-2022-testing}
Emmy Liu, Chenxuan Cui, Kenneth Zheng, and Graham Neubig.
\newblock Testing the ability of language models to interpret figurative language.
\newblock In Marine Carpuat, Marie-Catherine de~Marneffe, and Ivan~Vladimir Meza~Ruiz, editors, {\em Proceedings of the 2022 Conference of the North American Chapter of the Association for Computational Linguistics: Human Language Technologies}, pages 4437--4452, Seattle, United States, July 2022. Association for Computational Linguistics.

\bibitem{10534945}
Sze Ching~Evelyn Fung, Man~Fai Wong, and Chee~Wei Tan.
\newblock Chain-of-thoughts prompting with language models for accurate math problem-solving.
\newblock In {\em 2023 IEEE MIT Undergraduate Research Technology Conference (URTC)}, pages 1--5, 2023.

\bibitem{10.5555/3600270.3602070}
Jason Wei, Xuezhi Wang, Dale Schuurmans, Maarten Bosma, Brian Ichter, Fei Xia, Ed~H. Chi, Quoc~V. Le, and Denny Zhou.
\newblock Chain-of-thought prompting elicits reasoning in large language models.
\newblock In {\em Proceedings of the 36th International Conference on Neural Information Processing Systems}, NIPS '22, Red Hook, NY, USA, 2024. Curran Associates Inc.

\bibitem{DBLP:conf/aaai/WangHHXLLS24}
Lei Wang, Yi~Hu, Jiabang He, Xing Xu, Ning Liu, Hui Liu, and Heng~Tao Shen.
\newblock T-sciq: Teaching multimodal chain-of-thought reasoning via large language model signals for science question answering.
\newblock {\em Proceedings of the AAAI Conference on Artificial Intelligence}, 38(17):19162--19170, Mar. 2024.

\bibitem{Cohn2024ACP}
Clayton Cohn, Nicole~M. Hutchins, Tuan Le, and Gautam Biswas.
\newblock A chain-of-thought prompting approach with llms for evaluating students' formative assessment responses in science.
\newblock In {\em AAAI Conference on Artificial Intelligence}, 2024.

\bibitem{10337264}
Hangtian Zhao, Hakiz Yilahun, and Askar Hamdulla.
\newblock Pipeline chain-of-thought: A prompt method for large language model relation extraction.
\newblock In {\em 2023 International Conference on Asian Language Processing (IALP)}, pages 31--36, 2023.

\bibitem{DBLP:journals/corr/abs-2112-00114}
Maxwell~I. Nye, Anders~Johan Andreassen, Guy Gur{-}Ari, Henryk Michalewski, Jacob Austin, David Bieber, David Dohan, Aitor Lewkowycz, Maarten Bosma, David Luan, Charles Sutton, and Augustus Odena.
\newblock Show your work: Scratchpads for intermediate computation with language models.
\newblock {\em CoRR}, abs/2112.00114, 2021.

\bibitem{wei2022chain}
Jason Wei, Xuezhi Wang, Dale Schuurmans, Maarten Bosma, Fei Xia, Ed~Chi, Quoc~V Le, Denny Zhou, et~al.
\newblock Chain-of-thought prompting elicits reasoning in large language models.
\newblock {\em Advances in Neural Information Processing Systems}, 35:24824--24837, 2022.

\bibitem{NEURIPS2022_8bb0d291}
Takeshi Kojima, Shixiang~(Shane) Gu, Machel Reid, Yutaka Matsuo, and Yusuke Iwasawa.
\newblock Large language models are zero-shot reasoners.
\newblock In {\em Advances in Neural Information Processing Systems}, volume~35, pages 22199--22213, 2022.

\bibitem{ji2024chain}
Bin Ji, Huijun Liu, Mingzhe Du, and See-Kiong Ng.
\newblock Chain-of-thought improves text generation with citations in large language models.
\newblock In {\em Proceedings of the AAAI Conference on Artificial Intelligence}, volume~38, pages 18345--18353, 2024.

\bibitem{fei-etal-2023-reasoning}
Hao Fei, Bobo Li, Qian Liu, Lidong Bing, Fei Li, and Tat-Seng Chua.
\newblock Reasoning implicit sentiment with chain-of-thought prompting.
\newblock In Anna Rogers, Jordan Boyd-Graber, and Naoaki Okazaki, editors, {\em Proceedings of the 61st Annual Meeting of the Association for Computational Linguistics (Volume 2: Short Papers)}, pages 1171--1182, Toronto, Canada, July 2023. Association for Computational Linguistics.

\bibitem{GU2024107907}
Xu~Gu, Xiaoliang Chen, Peng Lu, Zonggen Li, Yajun Du, and Xianyong Li.
\newblock Agcvt-prompt for sentiment classification: Automatically generating chain of thought and verbalizer in prompt learning.
\newblock {\em Engineering Applications of Artificial Intelligence}, 132:107907, 2024.

\bibitem{zhang2023multicot}
Zhuosheng Zhang, Aston Zhang, Mu~Li, Hai Zhao, George Karypis, and Alex Smola.
\newblock Multimodal chain-of-thought reasoning in language models.
\newblock {\em arXiv preprint arXiv:2302.00923}, 2023.

\bibitem{zheng-etal-2024-enhancing-semantics}
Guangmin Zheng, Jin Wang, Xiaobing Zhou, and Xuejie Zhang.
\newblock Enhancing semantics in multimodal chain of thought via soft negative sampling.
\newblock In Nicoletta Calzolari, Min-Yen Kan, Veronique Hoste, Alessandro Lenci, Sakriani Sakti, and Nianwen Xue, editors, {\em Proceedings of the 2024 Joint International Conference on Computational Linguistics, Language Resources and Evaluation (LREC-COLING 2024)}, pages 6059--6076, Torino, Italia, May 2024. ELRA and ICCL.

\bibitem{he2024multi}
Liqi He, Zuchao Li, Xiantao Cai, and Ping Wang.
\newblock Multi-modal latent space learning for chain-of-thought reasoning in language models.
\newblock In {\em Proceedings of the AAAI Conference on Artificial Intelligence}, volume~38, pages 18180--18187, 2024.

\bibitem{Binz_2023}
Marcel Binz and Eric Schulz.
\newblock Using cognitive psychology to understand gpt-3.
\newblock {\em Proceedings of the National Academy of Sciences}, 120(6), February 2023.

\bibitem{Dasgupta2022LanguageMS}
Ishita Dasgupta, Andrew~Kyle Lampinen, Stephanie C.~Y. Chan, Antonia Creswell, Dharshan Kumaran, James~L. McClelland, and Felix Hill.
\newblock Language models show human-like content effects on reasoning.
\newblock {\em ArXiv}, abs/2207.07051, 2022.

\bibitem{Kao2014NonliteralUO}
Justine~T. Kao, Jean Wu, Leon Bergen, and Noah~D. Goodman.
\newblock Nonliteral understanding of number words.
\newblock {\em Proceedings of the National Academy of Sciences}, 111:12002 -- 12007, 2014.

\bibitem{prystawski2023psychologicallyinformed}
Ben Prystawski, Paul~H. Thibodeau, Christopher Potts, and Noah~D. Goodman.
\newblock Psychologically-informed chain-of-thought prompts for metaphor understanding in large language models, 2023.

\bibitem{liu2021investigating}
Risheng Liu, Jiaxin Gao, Jin Zhang, Deyu Meng, and Zhouchen Lin.
\newblock Investigating bi-level optimization for learning and vision from a unified perspective: A survey and beyond.
\newblock {\em IEEE Transactions on Pattern Analysis and Machine Intelligence}, 44(12):10045--10067, 2021.

\bibitem{dempe2020bilevel}
Stephan Dempe.
\newblock Bilevel optimization: theory, algorithms, applications and a bibliography.
\newblock {\em Bilevel optimization: advances and next challenges}, pages 581--672, 2020.

\bibitem{liu2022target}
Jinyuan Liu, Xin Fan, Zhanbo Huang, Guanyao Wu, Risheng Liu, Wei Zhong, and Zhongxuan Luo.
\newblock Target-aware dual adversarial learning and a multi-scenario multi-modality benchmark to fuse infrared and visible for object detection.
\newblock In {\em Proceedings of the IEEE/CVF conference on computer vision and pattern recognition}, pages 5802--5811, 2022.

\bibitem{liu2022task}
Risheng Liu, Long Ma, Xiaoming Yuan, Shangzhi Zeng, and Jin Zhang.
\newblock Task-oriented convex bilevel optimization with latent feasibility.
\newblock {\em IEEE Transactions on Image Processing}, 31:1190--1203, 2022.

\bibitem{forceville2017visual}
Charles Forceville.
\newblock Visual and multimodal metaphor in advertising: Cultural perspectives.
\newblock {\em Styles of communication}, 9(2), 2017.

\bibitem{zhang-etal-2023-multicmet}
Dongyu Zhang, Jingwei Yu, Senyuan Jin, Liang Yang, and Hongfei Lin.
\newblock {M}ulti{CMET}: A novel {C}hinese benchmark for understanding multimodal metaphor.
\newblock In Houda Bouamor, Juan Pino, and Kalika Bali, editors, {\em Findings of the Association for Computational Linguistics: EMNLP 2023}, pages 6141--6154, Singapore, December 2023. Association for Computational Linguistics.

\bibitem{cohen1960coefficient}
Jacob Cohen.
\newblock A coefficient of agreement for nominal scales.
\newblock {\em Educational and psychological measurement}, 20(1):37--46, 1960.

\bibitem{pennington2014glove}
Jeffrey Pennington, Richard Socher, and Christopher~D Manning.
\newblock Glove: Global vectors for word representation.
\newblock In {\em Proceedings of the 2014 conference on empirical methods in natural language processing (EMNLP)}, pages 1532--1543, 2014.

\bibitem{mcinnes2018umap}
Leland McInnes, John Healy, and James Melville.
\newblock Umap: Uniform manifold approximation and projection for dimension reduction.
\newblock {\em stat}, 1050:18, 2020.

\bibitem{chang2013missing}
Chun-Tuan Chang and Ching-Ting Yen.
\newblock Missing ingredients in metaphor advertising: The right formula of metaphor type, product type, and need for cognition.
\newblock {\em Journal of advertising}, 42(1):80--94, 2013.

\bibitem{ochs2015bilevel}
Peter Ochs, Ren{\'e} Ranftl, Thomas Brox, and Thomas Pock.
\newblock Bilevel optimization with nonsmooth lower level problems.
\newblock In {\em Scale Space and Variational Methods in Computer Vision: 5th International Conference, SSVM 2015, L{\`e}ge-Cap Ferret, France, May 31-June 4, 2015, Proceedings 5}, pages 654--665. Springer, 2015.

\bibitem{lewis-etal-2020-bart}
Mike Lewis, Yinhan Liu, Naman Goyal, Marjan Ghazvininejad, Abdelrahman Mohamed, Omer Levy, Veselin Stoyanov, and Luke Zettlemoyer.
\newblock {BART}: Denoising sequence-to-sequence pre-training for natural language generation, translation, and comprehension.
\newblock In Dan Jurafsky, Joyce Chai, Natalie Schluter, and Joel Tetreault, editors, {\em Proceedings of the 58th Annual Meeting of the Association for Computational Linguistics}, pages 7871--7880, Online, July 2020. Association for Computational Linguistics.

\bibitem{dosovitskiy2020vit}
Alexey Dosovitskiy, Lucas Beyer, Alexander Kolesnikov, Dirk Weissenborn, Xiaohua Zhai, Thomas Unterthiner, Mostafa Dehghani, Matthias Minderer, Georg Heigold, Sylvain Gelly, Jakob Uszkoreit, and Neil Houlsby.
\newblock An image is worth 16x16 words: Transformers for image recognition at scale.
\newblock {\em ICLR}, 2021.

\bibitem{paszke2019pytorch}
Adam Paszke, Sam Gross, Francisco Massa, Adam Lerer, James Bradbury, Gregory Chanan, Trevor Killeen, Zeming Lin, Natalia Gimelshein, Luca Antiga, et~al.
\newblock Pytorch: An imperative style, high-performance deep learning library.
\newblock {\em Advances in neural information processing systems}, 32, 2019.

\bibitem{shao2021cpt}
Yunfan Shao, Zhichao Geng, Yitao Liu, Junqi Dai, Hang Yan, Fei Yang, Zhe Li, Hujun Bao, and Xipeng Qiu.
\newblock Cpt: A pre-trained unbalanced transformer for both chinese language understanding and generation.
\newblock {\em Science China Information Sciences}, 67(5):1--13, 2024.

\bibitem{devlin-etal-2019-bert}
Jacob Devlin, Ming-Wei Chang, Kenton Lee, and Kristina Toutanova.
\newblock {BERT}: Pre-training of deep bidirectional transformers for language understanding.
\newblock In Jill Burstein, Christy Doran, and Thamar Solorio, editors, {\em Proceedings of the 2019 Conference of the North {A}merican Chapter of the Association for Computational Linguistics: Human Language Technologies, Volume 1 (Long and Short Papers)}, pages 4171--4186, Minneapolis, Minnesota, June 2019. Association for Computational Linguistics.

\bibitem{Qwen-VL}
Jinze Bai, Shuai Bai, Shusheng Yang, Shijie Wang, Sinan Tan, Peng Wang, Junyang Lin, Chang Zhou, and Jingren Zhou.
\newblock Qwen-vl: A frontier large vision-language model with versatile abilities.
\newblock {\em arXiv preprint arXiv:2308.12966}, 2023.

\bibitem{cheng2023evaluating}
Qinyuan Cheng, Tianxiang Sun, Wenwei Zhang, Siyin Wang, Xiangyang Liu, Mozhi Zhang, Junliang He, Mianqiu Huang, Zhangyue Yin, Kai Chen, and Xipeng Qiu.
\newblock Evaluating hallucinations in chinese large language models, 2023.

\bibitem{liu2024visual}
Haotian Liu, Chunyuan Li, Qingyang Wu, and Yong~Jae Lee.
\newblock Visual instruction tuning.
\newblock {\em Advances in neural information processing systems}, 36, 2024.

\bibitem{carion2020end}
Nicolas Carion, Francisco Massa, Gabriel Synnaeve, Nicolas Usunier, Alexander Kirillov, and Sergey Zagoruyko.
\newblock End-to-end object detection with transformers.
\newblock In {\em European conference on computer vision}, pages 213--229. Springer, 2020.

\bibitem{bert-score}
Tianyi Zhang, Varsha Kishore, Felix Wu, Kilian~Q. Weinberger, and Yoav Artzi.
\newblock Bertscore: Evaluating text generation with bert.
\newblock In {\em International Conference on Learning Representations}, 2020.

\end{thebibliography}

\end{document}